\documentclass[runningheads]{llncs}
\usepackage{silence}
\usepackage{xr-hyper}
\usepackage[hidelinks]{hyperref}
\usepackage{graphicx}
\usepackage{amsmath,amssymb} 
\usepackage{xcolor}
\usepackage{mathtools}
\usepackage{microtype}
\usepackage{enumitem}
\setlist{nosep} 
\setlist{itemsep=1pt, topsep=3pt}

\usepackage{booktabs}
\usepackage{cite} 
\usepackage{multirow} 
\usepackage{xspace}
\usepackage{xparse}
\usepackage{mathrsfs} 
\usepackage{tabularx}
\usepackage{algorithm} 
\usepackage[noend]{algpseudocode} 
\usepackage{bbm} 
\usepackage{bm}
\usepackage[percent]{overpic}
\usepackage{subcaption}
\usepackage{etoolbox}
\usepackage{tikz}
\usepackage[capitalise]{cleveref}
\crefname{algocf}{Alg.}{Algs.}
\Crefname{algocf}{Algorithm}{Algorithms}

\usetikzlibrary{tikzmark}
\usetikzlibrary{calc}

\errorcontextlines\maxdimen

\newcommand{\ALGtikzmarkcolor}{black}
\newcommand{\ALGtikzmarkextraindent}{4pt}
\newcommand{\ALGtikzmarkverticaloffsetstart}{-.5ex}
\newcommand{\ALGtikzmarkverticaloffsetend}{-.5ex}
\makeatletter
\newcounter{ALG@tikzmark@tempcnta}

\newcommand\ALG@tikzmark@start{%
    \global\let\ALG@tikzmark@last\ALG@tikzmark@starttext%
    \expandafter\edef\csname ALG@tikzmark@\theALG@nested\endcsname{\theALG@tikzmark@tempcnta}%
    \tikzmark{ALG@tikzmark@start@\csname ALG@tikzmark@\theALG@nested\endcsname}%
    \addtocounter{ALG@tikzmark@tempcnta}{1}%
}

\def\ALG@tikzmark@starttext{start}
\newcommand\ALG@tikzmark@end{%
    \ifx\ALG@tikzmark@last\ALG@tikzmark@starttext
    \else
        \tikzmark{ALG@tikzmark@end@\csname ALG@tikzmark@\theALG@nested\endcsname}%
        \tikz[overlay,remember picture] \draw[\ALGtikzmarkcolor] let \p{S}=($(pic cs:ALG@tikzmark@start@\csname ALG@tikzmark@\theALG@nested\endcsname)+(\ALGtikzmarkextraindent,\ALGtikzmarkverticaloffsetstart)$), \p{E}=($(pic cs:ALG@tikzmark@end@\csname ALG@tikzmark@\theALG@nested\endcsname)+(\ALGtikzmarkextraindent,\ALGtikzmarkverticaloffsetend)$) in (\x{S},\y{S})--(\x{S},\y{E});%
    \fi
    \gdef\ALG@tikzmark@last{end}%
}

\apptocmd{\ALG@beginblock}{\ALG@tikzmark@start}{}{\errmessage{failed to patch}}
\pretocmd{\ALG@endblock}{\ALG@tikzmark@end}{}{\errmessage{failed to patch}}
\makeatother
\newcommand*{\eg}{e.g.\@\xspace}
\newcommand*{\ie}{i.e.\@\xspace}
\DeclareMathOperator*{\expectationop}{\mathbb{E}}
\newcommand{\expectation}[2]{\ensuremath{\expectationop_{#1}\left[#2\right]}}

\newcommand{\nmax}{n_{\max{}}}

\newcommand{\intrange}[2]{\{#1\,..\,#2\}}
\newcommand{\etal}{et al.}
\renewcommand{\question}[1]{\paragraph{#1}}

\newcommand{\ccc}{c}

\newcommand{\ccs}{s}
\newcommand{\ccx}{x}
\newcommand{\cci}{i}
\newcommand{\ccj}{j}
\newcommand{\ccxi}{\ccx_\cci}
\newcommand{\ccxj}{\ccx_\ccj}

\newcommand{\ccsx}{X} 
\newcommand{\ccsxp}{X'} 

\newcommand{\ccsxpwith}[1]{\ccsxp \cup \{#1\}}
\newcommand{\ccrvx}{\ccsx}
\newcommand{\ccrvxp}{\ccsx'}
\newcommand{\ccrvs}{\ccs}
\newcommand{\ccmc}[3]{\Delta^{#2}_{#1}(#3)}

\DeclarePairedDelimiterX{\RoundBrackets}[1]{(}{)}{#1}
\newcommand{\ccf}{f}
\newcommand{\ccfms}{f^{\mathrm{ms}}}
\newcommand{\ccfc}{f_\ccc}

\makeatletter
\renewcommand\paragraph{%
  \@startsection{paragraph}{4}%
  {\z@}{-4\p@ \@plus -2\p@ \@minus -2\p@}{-0.5em \@plus -0.22em \@minus -0.1em}%
  {\normalfont\normalsize\bfseries}%
}
\makeatother

\begin{document}
\pagestyle{headings}
\mainmatter
\def\ACCV20SubNumber{30}  

\title{Play Fair: Frame Attributions in Video Models}
\titlerunning{Play Fair: Frame Attributions in Video Models}
\authorrunning{W. Price and D. Damen}

\author{Will Price\orcidID{0000-0003-2884-0290} \and Dima Damen\orcidID{0000-0001-8804-6238}}
\institute{\email{<first>.<last>@bristol.ac.uk}, University of Bristol, UK}

\maketitle

\begin{abstract}
In this paper, we introduce an attribution method for explaining action recognition models.
Such models fuse information from multiple frames within a video, through score aggregation or relational reasoning.
We break down a model's class score into the sum of contributions from each frame, \textit{fairly}.
Our method adapts an axiomatic solution to fair reward distribution in cooperative games, known as the Shapley value, for elements in a variable-length sequence, which we call the Element Shapley Value (ESV).
Critically, we propose a tractable approximation of ESV that scales linearly with the number of frames in the sequence.

We employ ESV to explain two action recognition models (TRN and TSN) on the fine-grained dataset Something-Something.
We offer detailed analysis of supporting/distracting frames, and the relationships of ESVs to the frame's position, class prediction, and sequence length.
We compare ESV to naive baselines and two commonly used feature attribution methods: Grad-CAM and Integrated-Gradients.
\end{abstract}

\section{Introduction}
\label{sec:introduction}
Progress in Action Recognition has seen remarkable gains in recent years thanks to architectural advances~\cite{zhou2018_TemporalRelationalReasoning,tran2018_CloserLookSpatiotemporal,hussein2019_TimeceptionComplexAction,girdhar2019_VideoActionTransformer,chen2018_MultifiberNetworksVideo,lin2019_TSMTemporalShift,feichtenhofer2019_SlowFastNetworksVideo} and large-scale datasets~\cite{abu-el-haija2016_YouTube8MLargeScaleVideo,kay2017_KineticsHumanAction,goyal2017_SomethingSomethingVideo,diba2019_LargeScaleHolistic,gu2018_AVAVideoDataset,damen2018_ScalingEgocentricVisionb}.
The task of action recognition is to classify the action depicted in a video, from a sequence of frames.
We address a question previously unexplored in action recognition:
given a video and a trained model, \textit{how much did each frame contribute to the model's output?}

\begin{figure}[t]
    \includegraphics[width=.58\textwidth]{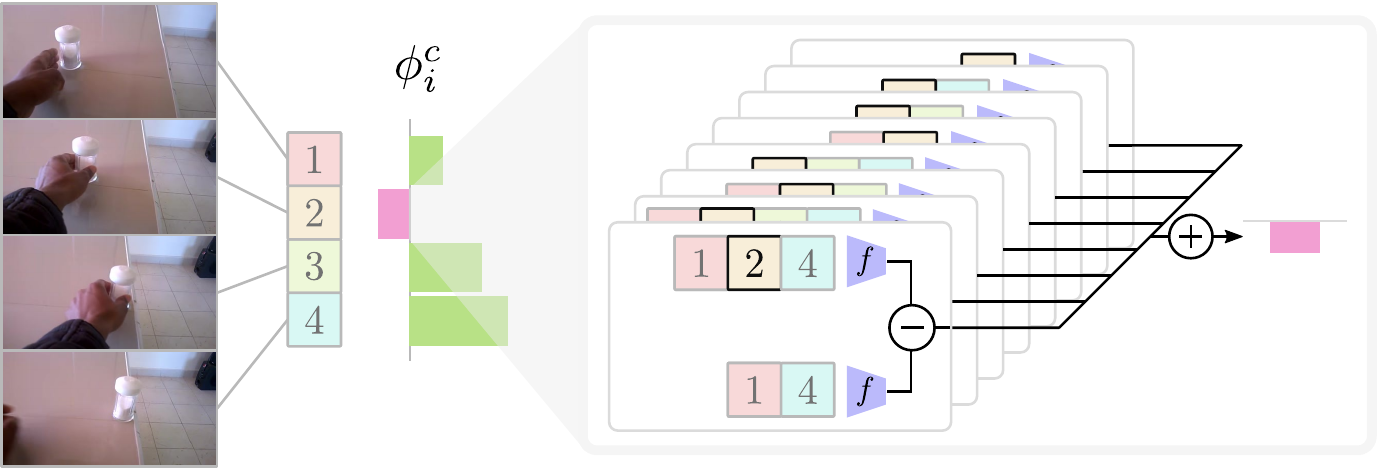}
    \includegraphics[width=.4\textwidth]{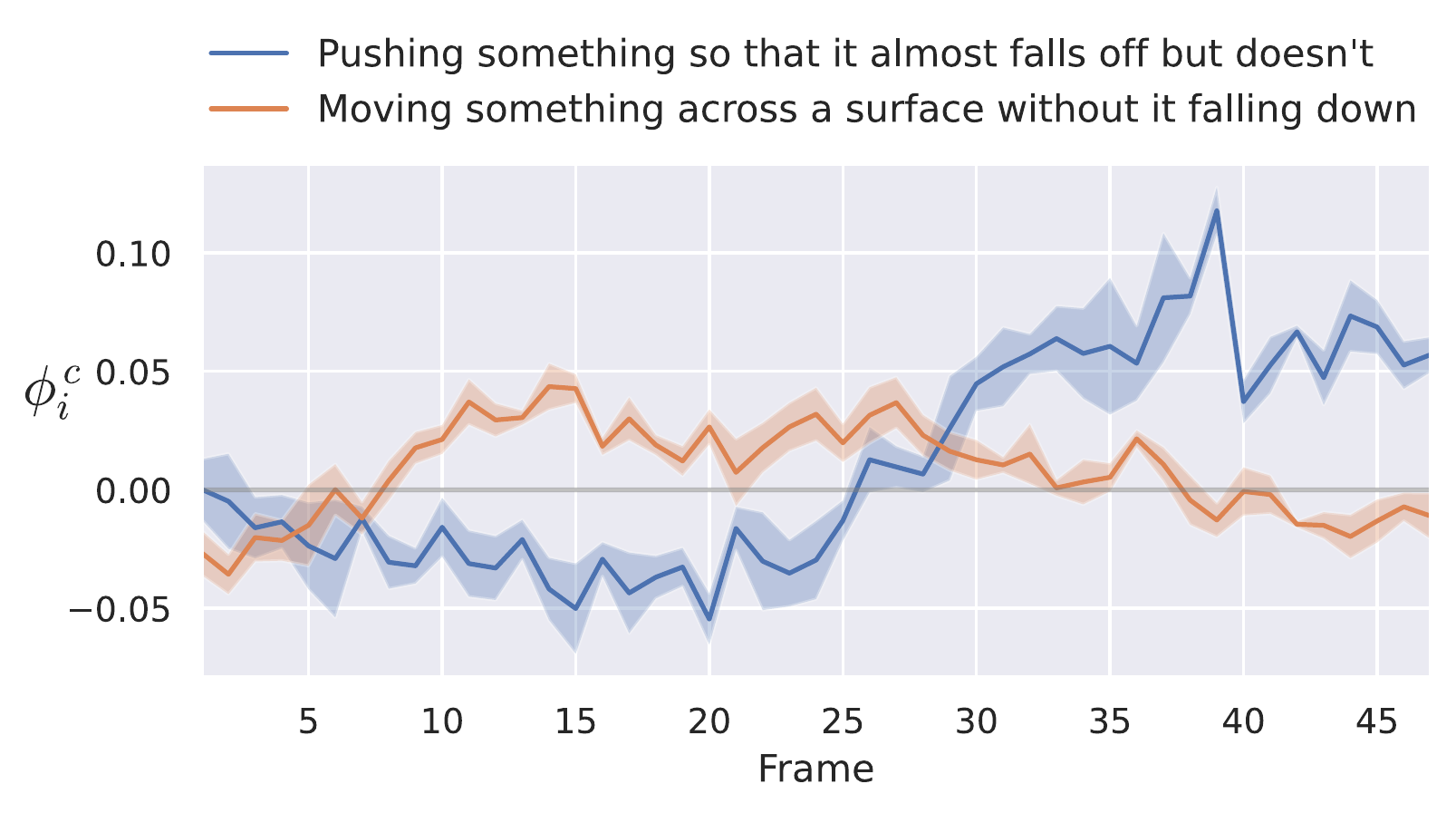}
    \caption{We assign $\phi^c_i$, the Element Shapley Value, to each frame $i$ in a video \textit{fairly}, given a model $f$ and a specific class $c$.
    We calculate differences in the class score for subsequences with and without the specified frame---\eg{} (1,\textbf{2},4) vs (1,4), and combine these for all subsequences (tiled boxes) to produce the element's attribution (\eg{} for frame 2). Class-specific $\phi_i^c$ are shown for two classes (right), the ground-truth class (in blue) and another class (in orange), highlighting positive/negative frame attributions.}
    \vspace*{-8pt}
    \label{fig:concept}
\end{figure}

Determining the contribution of each frame is similar to feature attribution, which assigns a value to each input feature, representing its weight to a scalar output in the model's response.
Note that this is distinct from feature selection which computes the value of the feature globally across a dataset, and not for a specific example.
Previous works in feature attribution~\cite{strumbelj2010_EfficientExplanationIndividual,strumbelj2014_Explainingpredictionmodels,lundberg2017_UnifiedApproachInterpreting,sundararajan2017_AxiomaticAttributionDeep,sundararajan2020_manyShapleyvalues} have taken inspiration from cooperative game theory 
The Shapley value~\cite{shapley1953_ValuenPersonGames} was proposed as an axiomatic solution to \textit{fairly} distributing a reward amongst the players in a cooperative game based on each player's contribution.
We offer the first attempt to integrate the Shapley axioms of fairness for video understanding---hence our paper's title `Play Fair'.

In contrast to general feature attribution, and similar approaches applied to images~\cite{selvaraju2017_GradCAMVisualExplanations,zhang2018_TopdownNeuralAttention,fong2019_UnderstandingDeepNetworks,bach2015_PixelWiseExplanationsNonLinear,fong2017_InterpretableExplanationsBlack,ribeiro2016_WhyShouldTrust}, our focus is on attributing elements (frames) in a time-varying sequence (video).
We depict our approach, \textit{Element Shapley Value} (ESV), in \cref{fig:concept}.
ESV assigns an attribution to each frame representing its contribution to the model output.
This attribution is computed from the change in the model output as the frame is added to different subsequences of the input video.
We compute \textit{class-specific} attributions (Fig.~\ref{fig:concept}~right).
Additionally, as attributions produced by ESV can be combined, we introduce \textit{class-contrastive} attributions to determine which frames contribute more to one class than another.

\noindent Our contributions can be summarised as follows:
\begin{enumerate}
  \item We introduce Element Shapley Value (ESV) to determine the contribution of each element in a variable-length sequence to a model's output\footnote{We focus on individual frames, but our approach naturally extends to determining attributions of video clips fed to a 3D CNN (\eg{}~\cite{carreira2017_QuoVadisAction}). As these models average clip scores, ESV can explain how much did each \textit{clip} contribute to the model's output.}.
 \item We propose and evaluate a tractable approximation to computing ESVs that linearly scales with the number of elements in the sequence.
  \item We calculate class-specific and class-contrastive ESVs for two action recognition models on the fine-grained dataset Something-Something.
  \item We demonstrate that some frames have a negative impact on the model's output. This deviates from the common misconception that utilising all frames in a sequence always improves performance.
\end{enumerate}
We next explain our approach to determining the contribution of each element in a sequence to a model's response,  
which we term \textit{element attribution}.
Related work is deferred until the end of the paper, to offer the best insight into how our work relates to feature attribution and approaches in video explanation.

\vspace*{-2pt}
\section{Element Attribution in Sequences}
\label{sec:element-attribution}
\vspace*{-4pt}
In this section, we introduce the Shapley value as an axiomatic solution to attribution.
We discuss the limitations of feature attribution methods, showcasing how these can be overcome when attributing elements in sequences.
We then introduce our approach to tractably computing ESVs for variable-length sequences.

\vspace*{-6pt}
\subsection{Element Attribution and the Shapley Value}
\label{sec:element-attribution:feature-attribution}
\vspace*{-2pt}

To answer our motivational question, \textit{how much did each frame contribute to the model's output?}, we introduce \textit{element attribution} as the task of determining the contribution $\phi_\cci$ of an element $\ccxi$ in a sequence $\ccsx = (\ccxi)^n_{i=1}$ to the output of the model $\ccf(\ccsx)$.
For a classification model, where the output is a vector of class scores $\ccf(\ccsx) \in \mathbb{R}^C$,
we obtain the contribution $\phi_i^c$ for each element $\ccxi$ to the score $\ccfc(\ccsx)$ of any class $c$.

Element attribution can be viewed as special case of grouped-feature attribution, where an element is represented by a group of features and elements are ordered by their observation time.
Of the previously-proposed feature attribution methods,
\textit{additive} ones~\cite{lundberg2017_UnifiedApproachInterpreting} have a desirable trait where the sum of attributions equals the model output, up to a constant additive bias $b_c$ specific to $\ccf$:
\begin{equation} \label{eq:base}
\vspace*{-6pt}
\ccfc(\ccsx) = b_c + \sum_{\ccxi \in \ccsx} \phi_i^c.
\end{equation}

Dividing the model's output should be done \textit{fairly}, reflecting the element's actual contribution to the model's output.
A natural way of measuring the contribution of an element $\ccxi$, is to consider the change in the model's response when it is added%
\footnote{
We adapt set notation to sequences using subset ($\subset$) and union ($\cup$) operations to form new subsequences. Elements preserve the same ordering as in the full sequence.
}
to a subsequence $\ccsxp \subset \ccsx$, where $\ccsxp$ does not contain $x_i$:
\begin{equation}
\label{eq:delta}
\ccmc{\cci}{\ccc}{\ccsxp} = \ccfc(\ccsxpwith{\ccxi}) - \ccfc(\ccsxp),
\end{equation}
this is known as the \textit{marginal contribution} of $\ccxi$ on $\ccsxp$.
Note when $\ccmc{\cci}{\ccc}{\ccsxp}$ is positive, the addition of $\ccxi$ increases the response of the model for class $c$ and when negative, decreases it.
However, $\ccmc{\cci}{\ccc}{\ccsxp}$ will differ depending on the subsequence $\ccsxp$.
As such, it is necessary to consider the marginal contribution of $\ccxi$ on all subsequences it can join to calculate element attributions.
This has the benefit of capturing the effects of resampling the sequence at all possible rates.

Lloyd Shapley (1953) defined  a set of axioms that $\phi^\ccc_\cci$ should satisfy if it is to be considered \textit{fair}~\cite{shapley1953_ValuenPersonGames} (later refined by Young~\cite{young1985_Monotonicsolutionscooperative} in 1985), we explain these in the context of element attribution:

\paragraph{A1: Efficiency.}
The sum of element attributions should equal the output of the model on the input $\ccsx$, minus the output of the model on the empty sequence $\emptyset$:
\begin{equation}
\sum_{\ccxi \in \ccsx} \phi_\cci^c = \ccfc(\ccsx) - \ccfc(\emptyset).
\end{equation}
We call this difference, $\ccfc'(\ccsx) = \ccfc(\ccsx) - \ccfc(\emptyset)$, the \textit{evidential score} of $\ccsx$ w.r.t to class $\ccc$, as it is the difference in class score when $\ccsx$ is observed over observing nothing.
 We use the empirical class distribution of the training set as $\ccf_c(\emptyset)$.

\paragraph{A2: Symmetry.}
Any pair of elements $\ccxi$ and $\ccxj$ should be assigned equal contributions, $\phi^\ccc_\cci = \phi^\ccc_\ccj$, if $\ccfc(\ccsxpwith{\ccxi}) = \ccfc(\ccsxpwith{\ccxj})$ for all $\ccsxp \subseteq \ccsx \setminus \{\ccxi, \ccxj\}$.
\paragraph{A3: Monotonicty.}
For any pair of classes $\ccc_1, \ccc_2$, if
$\ccmc{\cci}{\ccc_1}{\ccsxp} \geq \ccmc{\cci}{\ccc_2}{\ccsxp}$
for all subsequences $\ccsxp \subseteq \ccsx \setminus \{\ccxi\}$ then
$\phi^{\ccc_1}_i \geq \phi^{\ccc_2}_i$.

There is a unique formulation of $\phi^c_i$ satisfying the above axioms, known as the Shapley value, which consequently assigns attributions \textit{fairly}. The Shapley value~\cite{shapley1953_ValuenPersonGames} was originally proposed as a solution to the problem of fairly distributing a reward amongst players in a cooperative game.
A cooperative game is one where a group of players known as a \textit{coalition} collaborate to earn some reward.
Players are treated as sets, as opposed to sequences.
By analogy, each element in the sequence acts as a player and subsequences act as ordered coalitions.
Accordingly, we refer to our element attribution method, that satisfies the Shapley axioms, as the Element Shapley Value (ESV).
This is the weighted average of the marginal contributions of $\ccxi$ to all subsequences  $\ccsxp \subseteq \ccsx \setminus \{\ccxi\}$ it can join:
\begin{equation}
\label{eq:shapley-value:delta-form}
\phi_i^c = \sum_{\ccsxp \subseteq \ccsx \setminus \{\ccxi\}} w(\ccsxp) \ccmc{\cci}{\ccc}{\ccsxp} \qquad
w(\ccsxp) = \frac{(|\ccsx| - |\ccsxp| - 1)!|\ccsxp|!}{|\ccsx|!}\,.
\end{equation}
where $w(\ccsxp)$ can be interpreted as the probability of drawing a subsequence $\ccsxp$ from $\ccsx \setminus \{\ccxi\}$, considering each way of forming $\ccsx$ by adding an element at a time starting from $\emptyset$, going via $\ccsxp$ and $\ccsxpwith{\ccxi}$, as equally likely.
Consequently, $\phi^c_i$ can be rewritten as the expectation over a random variable $\ccrvxp$ with sample space $2^{\ccsx \setminus \{\ccxi\}}$ and probability mass function $w(\ccrvxp)$ (proof in \cref{appendix:shapley-value-forms:single-expectation}):
\begin{equation}
\phi_i^c = \expectation{\ccrvxp}{\ccmc{\cci}{\ccc}{\ccrvxp}},
\label{eq:shapley-value:delta-form-expectation}
\end{equation}
which can be reformulated again in terms of the expected model response on the sequences before and after $x_i$ is added to $\ccsxp$:
\begin{equation}
  \phi_i^c = \expectation{\ccrvxp}{\ccfc(\ccrvxp \cup \{\ccxi\})} - \expectation{\ccrvxp}{\ccfc(\ccrvxp)}\,.
  \label{eq:shapley-value:expectation-split}
\end{equation}
This is an easier form to compute, as we don't have to measure the marginal contributions of each element on each subsequence, but instead measure the expected model score on subsequences containing $\ccxi$ and those not containing $\ccxi$.
We take one final step towards the form that we will actually use to compute element attributions, expanding \cref{eq:shapley-value:expectation-split} via the law of total expectation to
\begin{equation}
  \label{eq:shapley-value:expectation-split-multi-scale}
  \phi_i^c = \expectation{\ccrvs}{\expectation{\ccrvxp | \ccrvs}{\ccfc(\ccrvxp \cup \{\ccxi\})} - \expectation{\ccrvxp | \ccrvs}{\ccfc(\ccrvxp)}},
\end{equation}
where the subsequence length $\ccrvs$ ranges from $0$ to $|\ccsx| - 1$ with equal probability, and the random variable $\ccrvxp$ conditioned on $\ccrvs$ takes on values from $\{\ccsxp \subseteq \ccsx \setminus \{\ccxi\} : |\ccsxp| = \ccrvs\}$ with equal probability (proof in \cref{appendix:shapley-value-forms:conditional-expectation}).
This allows us to compute $\phi^c_i$ by first considering the expected marginal contributions at each scale~$\ccs$ and then combining these across scales.

\subsection{Element Attribution in Variable-Length Sequences}
\label{sec:ablating-elements}
As shown in \cref{eq:shapley-value:delta-form}, computing element attributions requires evaluating the model on different subsequences.
We first consider approaches used to evaluate a model on feature subsets, then the options we have in element attribution.

\paragraph{Feature Ablation.}
Evaluating a model on feature subsets is challenging as models rarely support the notion of a `missing feature'.
Two approaches exist: re-training the model on all combinations of features~\cite{strumbelj2009_Explaininginstanceclassifications} or substituting missing features with those from a reference~\cite{strumbelj2010_EfficientExplanationIndividual,ribeiro2016_WhyShouldTrust,lundberg2017_UnifiedApproachInterpreting,fong2017_InterpretableExplanationsBlack,fong2019_UnderstandingDeepNetworks,manttari2020_Interpretingvideofeatures,li2020_ComprehensiveStudyVisual}, but both approaches have limitations.
Retraining is computationally infeasible for more than a handful of features, and the choice of reference in feature substitution has a significant impact on the resulting attribution values~\cite{sturmfels2020_VisualizingImpactFeature,sundararajan2020_manyShapleyvalues}.

\paragraph{Element Ablation.}
Training an exponential number of models for each subsequence is infeasible for large models and thus ruled out.
We could substitute frames by a \textit{reference}, \eg{} the mean element from the training set, or the nearest element still present, amongst other reasonable alternatives.
However, similar to feature substitution~\cite{sturmfels2020_VisualizingImpactFeature}, the choice of reference will have a large impact on the resulting attribution values.
Feeding sequences that are out-of-distribution (such as one created by duplicating frames) to an action recognition model can result in uncharacteristic model responses~\cite{huang2018_WhatMakesVideo}.
We instead propose a more attractive option: utilising models that support variable-length input sequences.

\paragraph{Supporting Variable Length Inputs.}
When the model to be explained does not support variable-length inputs, we take inspiration from multi-scale models (\eg{} TRN~\cite{zhou2018_TemporalRelationalReasoning}) to build a model $\ccfms$ capable of operating over variable-length sequences.
Let $\ccf^\ccs(\ccsx)$ be a fixed-scale model which takes a sequence of length~$\ccs$ as input.
We construct a set of models operating at different scales $\{\ccf^\ccs\}_{\ccs=1}^{\nmax{}}$, up to a maximum subsequence length $\nmax{}$. We then combine these, such that each scale contributes equally to the overall output:
\begin{equation}
  \label{eq:multi-scale-model}
  \ccfms(\ccsx) = \expectation{\ccrvs}{\expectation{\ccsxp | \ccrvs}{\ccf^{\ccrvs}(\ccsxp)}}\,,
\end{equation}
where $\ccrvs$ is a random variable over subsequence lengths $\intrange{1}{\min(|\ccsx|, \nmax)}$, all equally likely, and $\ccsxp$ is a random variable over subsequences of size $\ccrvs$.
This has similarities to the retraining approach~\cite{strumbelj2009_Explaininginstanceclassifications}, however we can leverage the homogeneity of our input to reduce the number of models from $\mathcal{O}(2^{|\ccsx|})$ (one for each possible subsequence) to $\mathcal{O}(\nmax{})$. Another contrast is that the same models are used in inference and element attribution, unlike the work in~\cite{strumbelj2009_Explaininginstanceclassifications}.

\subsection{Tractable Approach to Element Shapley Values}
\label{sec:approximation}

We have now defined the Element Shapley Value (ESV), as well as how these can be calculated using models supporting variable-length inputs.
We next show how we can tractability compute the ESVs  for \textit{all elements in a variable-length sequence in parallel} culminating in \cref{alg:shapley-value-computation}.

\paragraph{Bottom-up Computation.}
We compute the model's output for each subsequence once by deriving a recurrence relation between $\ccfms(\ccsxp)$ and $\ccfms(\ccsx)$, where $\ccsxp \subset \ccsx$, $|\ccsx| = n$, and $|\ccsxp| = n - 1$ (proof in \cref{appendix:recursive-definition-of-aggregation-module}),
\begin{equation}
  \label{eq:multi-scale-model:recurrent}
  \ccfms(\ccsx) =
  \begin{cases}
      \ccf^1(\ccsx) & n = 1\\
      \frac{1}{n}\left[ \ccf^{n}(\ccsx) + (n - 1)\expectation{\ccrvxp|\ccrvs = n - 1}{\ccfms(\ccrvxp)}\right] & n \leq \nmax \\
      \expectation{\ccrvxp|\ccrvs = n - 1}{\ccfms(\ccrvxp)}, & n > \nmax\\
  \end{cases}
\end{equation}
To compute $\ccfms(\ccsx)$, we start with subsequences of one element, moving up one scale at a time by combining results according to \cref{eq:multi-scale-model:recurrent}, obtaining $\ccfms(\ccsxp)$ for all $\ccsxp \subseteq \ccsx$ in the process.
To compute ESVs simultaneously, we compute the expected model scores on subsequences with/without each element at each scale.
We then combine these to obtain all elements' ESVs as in \cref{eq:shapley-value:expectation-split-multi-scale}.

\paragraph{Sampling Subsequences.}
The bottom-up computation improves the efficiency, but doesn't deal with the exponential number of subsequences.
We apply a sampling approximation to the expectations over subsequences in the definition of $\ccfms$ (\cref{eq:multi-scale-model:recurrent}) and the ESV (\cref{eq:shapley-value:expectation-split-multi-scale}).
This is similar to the Monte-Carlo approach used in prior applications of the Shapley value~\cite{strumbelj2010_EfficientExplanationIndividual,strumbelj2014_Explainingpredictionmodels} but is more sample efficient.
Our sampling approach aims to maximise the number of subsequence relationships across scales to best approximate $\ccfms$.
Given a sample of subsequences $\mathcal{X}^\ccs$ at scale $\ccs$, we grow subsequences by one element to obtain subsequences at the next scale,
We then form a pool of all possible candidate subsequences of length $\ccs + 1$:
\begin{equation}
\label{eq:grow-subsequences}
    \mathcal{C}^{\ccs + 1} = \bigcup_{\mathcal{X}^\ccs_j \in \mathcal{X}^\ccs}\{\mathcal{X}^\ccs_j \cup \{x\}: \, x \in X \setminus \mathcal{X}^\ccs_j\}\,,
\end{equation}
from which we sample at most $m$ subsequences to construct $\mathcal{X}^{\ccs + 1}$.
We start our sampling approach with all single element subsequences: $\mathcal{X}^1 = \{(\ccxi)\}_{i=1}^n$.

\paragraph{Tying it All Together.}
Combining the above techniques, we present our approach in \cref{alg:shapley-value-computation} for computing ESVs.
When the number of sampled subsequences, $m$, is chosen to be $\max_k {\binom{|\ccsx|}{k}}$, it computes the \textit{exact} ESVs $\phi^c_i$ for $\ccfms(\ccsx)$.
When $m$ is less than this, the algorithm computes the approximate ESVs $\hat{\phi}^c_i$.
We repeat the inner loop that computes the marginal contributions across all scales a number of iterations to improve the accuracy of the approximation.
\begin{algorithm}[t!]
  \caption{Element Shapley Value (ESV) computation using $\ccfms$}
  \label{alg:shapley-value-computation}
  \textbf{Input}: A sequence $\ccsx = (\ccx_i)^n_{i=1}: \mathbb{R}^{n \times D}$,
    Single-scale models $\{\ccf^\ccs: \mathbb{R}^{\ccs\times D} \rightarrow \mathbb{R}^{C}\}_{\ccs = 1}^{\nmax{}}$ comprising $\ccfms$, and a class $c$ to explain.\\
  \textbf{Output}: Element Shapley Values $\phi^\ccc_\cci$ for all $\ccxi \in \ccsx$\\
  \textbf{Intermediates:} \\
  $\mathcal{X}^\ccs_j$: A subsequence of $\ccsx$ of length $\ccs$.\\
  $\mathcal{S}^\ccs_\cci, \bar{\mathcal{S}}^\ccs_\cci$: Sum of scores over sequences that contain/don't contain element $\ccxi$ at scale $\ccs$.\\
  $\mathcal{N}^\ccs_\cci, \bar{\mathcal{N}}^\ccs_\cci$: Number of sequences that contain/don't contain element $\ccxi$ at scale $\ccs$.\\
  $\mathcal{F}^\ccs_j$: Results of $\ccfms_\ccc$ on the subsequence $\mathcal{X}^\ccs_j$.
  \begin{algorithmic}[1]
  \State $\bar{\mathcal{S}}^0_i \leftarrow f_c(\emptyset)$,\; $\bar{\mathcal{N}}^0_i \leftarrow 1,\; \mathcal{X}^0 \leftarrow \{\{\}\}$

  \For {iteration from 0 to max iterations}
    \For {scale $s$ from 1 to $n$}
       \State $\mathcal{C}^\ccs$: Form extended subsequence candidate pool  according to \cref{eq:grow-subsequences}.
       \State $\mathcal{X}^\ccs \leftarrow$ Randomly select $\min(m, |\mathcal{C}^\ccs|)$ subsequences from $\mathcal{C}^\ccs$.
       \State $Z_j \leftarrow \sum_{k} \mathbbm{1}[\mathcal{X}^{\ccs - 1}_k \subset \mathcal{X}^\ccs_j]$
       \If{$s = 1$}
        \State $\mathcal{F}^1_j \leftarrow  \ccf^1_\ccc(\mathcal{X}^\ccs_j)$
       \ElsIf{$s \leq \nmax{}$}
        \State $\mathcal{F}^\ccs_j \leftarrow \frac{1}{\ccs} \left( \ccf^\ccs_\ccc(\mathcal{X}^\ccs_j) + (\ccs - 1) \frac{1}{Z_j} \sum_{k} \mathbbm{1}[\mathcal{X}^{\ccs - 1}_k \subset \mathcal{X}^\ccs_j] \mathcal{F}^{\ccs - 1}_{k} \right)$
       \Else
        \State $\mathcal{F}^\ccs_j \leftarrow \frac{\ccs - 1}{\ccs} \frac{1}{Z_j}\sum_{k} \mathbbm{1}[\mathcal{X}^{\ccs - 1}_k \subset \mathcal{X}^\ccs_j] \mathcal{F}^{\ccs - 1}_{k} $
       \EndIf
       \State $\mathcal{S}^\ccs_\cci \leftarrow \mathcal{S}^\ccs_\cci + \sum_j \mathbbm 1[\ccxi \in \mathcal{X}^\ccs_j] \mathcal{F}^\ccs_j, \quad \bar{\mathcal{S}}^\ccs_\cci \leftarrow \bar{\mathcal{S}}^\ccs_\cci + \sum_j \mathbbm 1[\ccxi \not\in \mathcal{X}^\ccs_j] \mathcal{F}^\ccs_j$
       \State $\mathcal{N}^\ccs_\cci \leftarrow \mathcal{N}^\ccs_\cci + \sum_j \mathbbm 1[\ccxi \in \mathcal{X}^\ccs_j], \quad \bar{\mathcal{N}}^\ccs_\cci \leftarrow \bar{\mathcal{N}}^\ccs_\cci + \sum_j \mathbbm 1[\ccxi \not\in \mathcal{X}^\ccs_j]$
       \EndFor
  \EndFor
  \State $\phi^\ccc_\cci \leftarrow \frac{1}{|\ccsx|} \sum_{\ccs=1}^{|X|} \mathcal{S}^\ccs_\cci/\mathcal{N}^\ccs_\cci  - \bar{\mathcal{S}}^{\ccs - 1}_\cci / \bar{\mathcal{N}}^{\ccs - 1}_\cci$
  \end{algorithmic}
\end{algorithm}

\paragraph{Additional Definitions.}
Elements whose $\phi_i^c > 0$ are termed \textit{supporting} elements for class $c$, otherwise, if $\phi_i^c \le 0$, they are termed \textit{distracting} elements.

A common question when diagnosing a model is to understand why it classified a sequence incorrectly.
We utilise the linear property of ESVs to compute a class-contrastive ESV $\delta_i$. This is the ESV for the class-contrastive (cc) model $f_{cc}(X) = f_{gt}(X)-f_{pt}(X)$, where $gt$ is the ground-truth class and $pt$ is the incorrectly predicted class. The class-constrastive ESV can thus be computed as: $\delta_i = \phi_i^{gt-pt} = \phi_i^{gt} - \phi_i^{pt}$.
When $\delta_i > 0$, the element contributes more to the ground-truth class than the predicted class.

\paragraph{Limitations.}
We foresee two limitations to our current implementation of ESVs:
(i) when the model's output for a given class is too small $\epsilon$, frame-level ESVs are uninformative; 
(ii)~when frame-level features are similar for multiple frames in the input sequence, our implementation calculates ESV for each independently, and does not benefit from feature similarities.

\section{Experiments}
\label{sec:experiments}

We have now explained how ESVs can be computed in a tractable manner for all elements in a variable-length sequence.
While our approach is applicable to sequences of any type of data, in this paper we focus on frames in video sequences.
We compare ESV against baselines and two other commonly used feature attribution methods in a frame ablation experiment.
We show that the choice of attribution method can have a big impact on the frames' attribution values.
We then analyse exact ESVs across a range of facets demonstrating how they can be used to understand models' behaviour.
Finally, we evaluate our proposed approximation showing that we can scale up our method to compute ESVs for all frames in variable-length videos.

\paragraph{Experimental Setup.}
\label{sec:experiments:setup}
Our experiments are conducted on the validation set of the large-scale Something-Something v2 dataset~\cite{goyal2017_SomethingSomethingVideo}, frequently used to probe video models ~\cite{dwibedi2018_TemporalReasoningVideos,manttari2020_Interpretingvideofeatures,zhou2018_TemporalRelationalReasoning}  due to its fine-grained nature, large number of classes~(174), and the temporal characteristics of most of its classes \eg{} ``Pulling [...] from behind of [...]''.
Unless otherwise stated, $\phi^c_i$ is calculated for the ground-truth class of the example video.
We implement $\ccfms$ using single hidden-layer MLPs atop of extracted features with $\nmax{} = 8$.

ESV is best applied to models where frame level features are extracted and combined through a temporal module.
In this case, only the temporal module is evaluated to compute the ESVs.
We primarily explain one commonly used action recognition model:
Temporal Relational Network
(TRN)~\cite{zhou2018_TemporalRelationalReasoning}. However, we also use
ESVs to explain Temporal Segment Network
(TSN)~\cite{wang2016_TemporalSegmentNetworks}, and provide open-source
code for explaining other models\footnote{\url{https://github.com/willprice/play-fair/}}.
Further implementation and training details can be found in \cref{appendix:experimental-setup} in addition to how we compute ESVs for TSN.

\subsection{Analysing Element Shapley Values}
\label{sec:experiments:shapley-values}
Our analysis is provided by answering questions that give new insights into how individual frames contribute to the model's output.

\question{How do ESVs differ from other attribution methods?}
\begin{figure}[t]
  \centering
  \includegraphics[width=\textwidth]{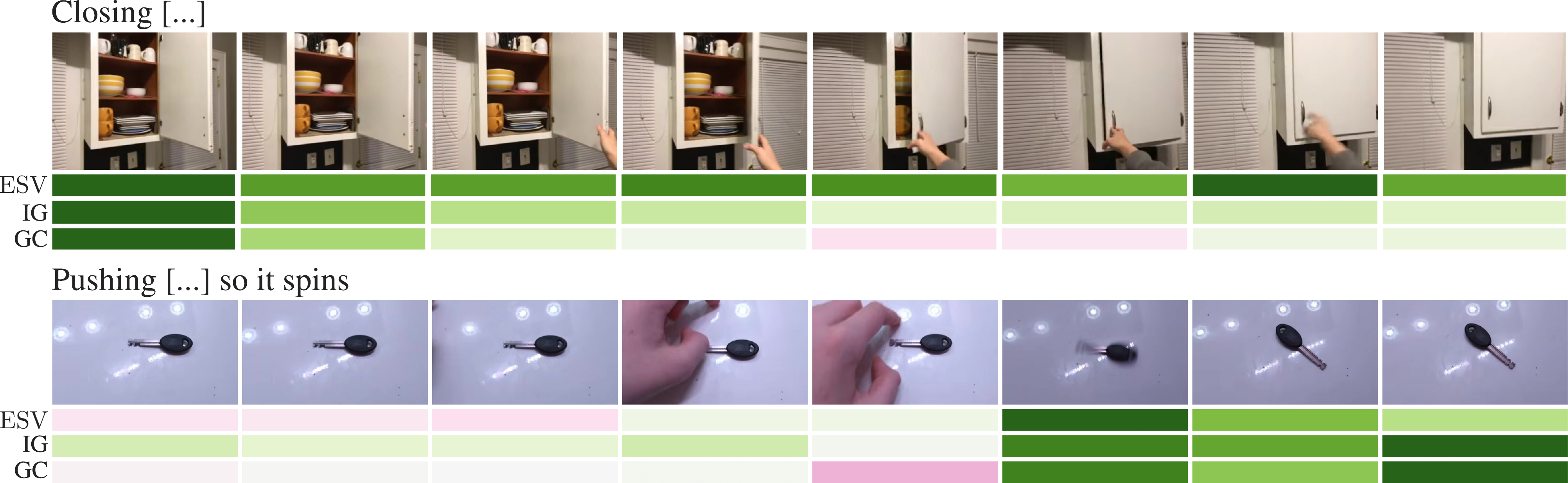}
  \vspace{-1.7em}
  \caption{Element attributions for two example sequences comparing our method (ESV) to Integrated Gradients (IG), and Grad-CAM (GC).
  \protect\includegraphics[height=.7em]{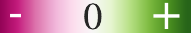}
  }
  \label{fig:method-comparison-qualitative}
\end{figure}

We compare ESV against Integrated Gradients (IG)~\cite{sundararajan2017_AxiomaticAttributionDeep} and Grad-CAM\footnote{We remove the ReLU from the formulation to produce signed attributions.} (GC)~\cite{selvaraju2017_GradCAMVisualExplanations}.
These methods produce attributions for each feature, so to obtain element attributions, we marginalise the spatial and channel dimensions by averaging.
We chose GC as the most-commonly used approach in explaining networks for video-understanding (\eg{}~\cite{chattopadhay2018_GradCAMGeneralizedGradientBased,doughty2018_WhoBetterWho,goyal2018_Evaluatingvisualcommona,srinivasan2017_Interpretablehumanaction}) and IG as an axiomatic approach based on Aumann-Shapley values~\cite{aumann1974_ValuesNonatomicGames}, an extension of the Shapley value to infinite player games.
IG computes the integral of gradients along a path from a reference to the input being explained (we use the mean frame feature as our reference).
We use the public implementation of these approaches from Captum~\cite{captum2019github}.

We first present a qualitative comparison on two sequences in \cref{fig:method-comparison-qualitative} demonstrating disagreement amongst the methods in their element attributions.
The top example (``Closing [...]'') shows that our method (ESV) highlights the first frame showing the open cupboard as well as the frame where the cupboard door has just been closed as the most important. Both IG and GC only highlight the first frame, missing the action completion, with GC considering the completion as marginally distracting.
In the bottom example, ESV assigns the first three frames negative attributions, with positive attributions restricted to frames after the hand appears. ESV considers the frame with the visibly spinning key as most important while IG and SG highlight the last frame instead.
\begin{figure}[t]
  \centering
  \begin{minipage}[t]{\textwidth}
    \vspace{0pt}  
    \centering
    \begin{overpic}[width=\textwidth]{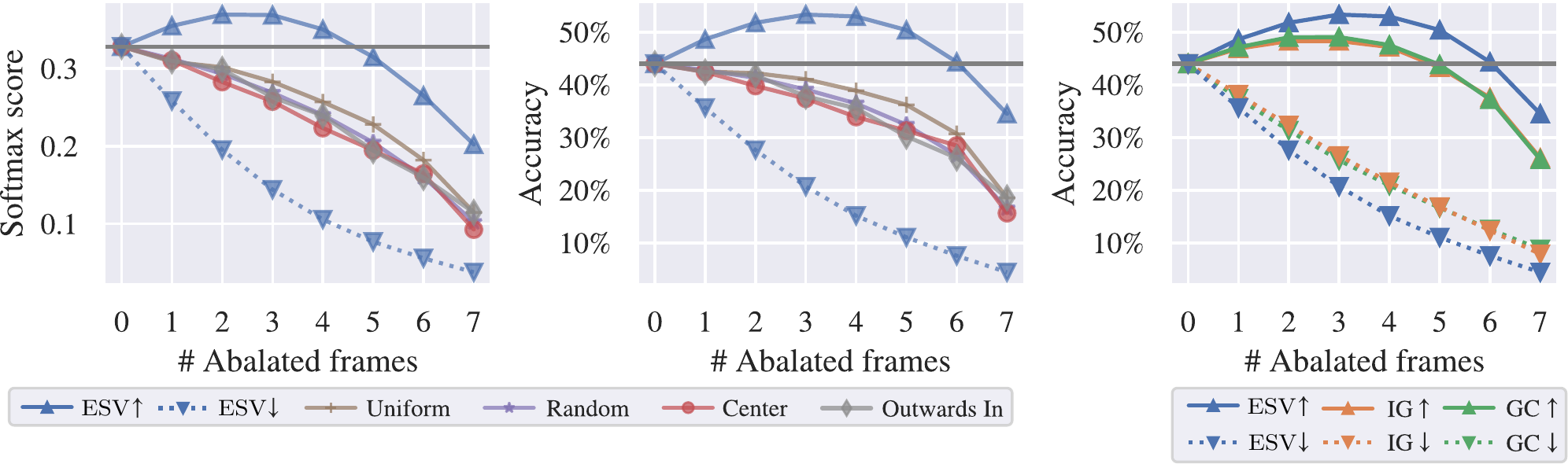}
    \put (0, 6) {(a)}
    \put (33.5,6) {(b)}
    \put (67.5,6) {(c)}
    \end{overpic}
    \phantomsubcaption
    \label{fig:frame-ablation:score}
    \phantomsubcaption
    \label{fig:frame-ablation:accuracy}
    \phantomsubcaption
    \label{fig:method-comparison-frame-ablation}
  \end{minipage}
  \vspace{-2em}
  \caption{
    TRN class score and accuracy after iteratively discarding frames in order of their attribution rank (ascending $\blacktriangle$ vs descending $\blacktriangledown$). We compare our method~(ESV) to baselines (a,b) and two alternate attribution methods (c): GradCam (GC) and Integrated Gradients (IG). We keep figures (b) and (c) separate for readability.
    Removing frames with the highest ESV first causes the quickest degradation, whilst removing frames with the lowest ESV improves performance by avoiding distractor frames.
  }
  \label{fig:frame-ablation}
\end{figure}

We compute the Pearson correlation between the attribution values produced by each pair of methods on the validation set.
For ESV$\times$IG, ESV$\times$GC and IG$\times$GC we get 0.60, 0.56 and 0.81 respectively for TRN compared to 0.85, 0.89 and 0.82 for TSN.
We note that agreement differs per model. IG is more similar to GC when analysing TRN, but more similar to ESV when analysing TSN model attributions.
Critically, we believe our fairness axioms and avoidance of out-of-distribution inputs make ESV a more founded technique for element attribution as we will demonstrate next.

\question{How does performance change when we discard frames according to their attributions?}
For TRN, we iteratively discard frames from the input sequence in descending/ascending order of their attributions.
We compare this approach to four baselines:
discarding frames from the center frame outwards, from the edges of the sequence inwards, uniformly, and randomly.
\Cref{fig:frame-ablation:score,fig:frame-ablation:accuracy} report results of this investigation on the full validation set uniformly sampling 8 frames from each video.
Whilst a boost in performance is expected when removing frames with the lowest attribution value, since this approach has privileged knowledge of the ground-truth class, these figures show that
(i) on average, 4 frames in an 8-frame sequence negatively influence the class score, and
(ii) it is possible to maintain the model's accuracy whilst discarding the majority~(6/8) of the frames.

\begin{figure}[t]
  \centering
  \begin{minipage}[t]{\textwidth}
      \begin{overpic}[width=\textwidth]{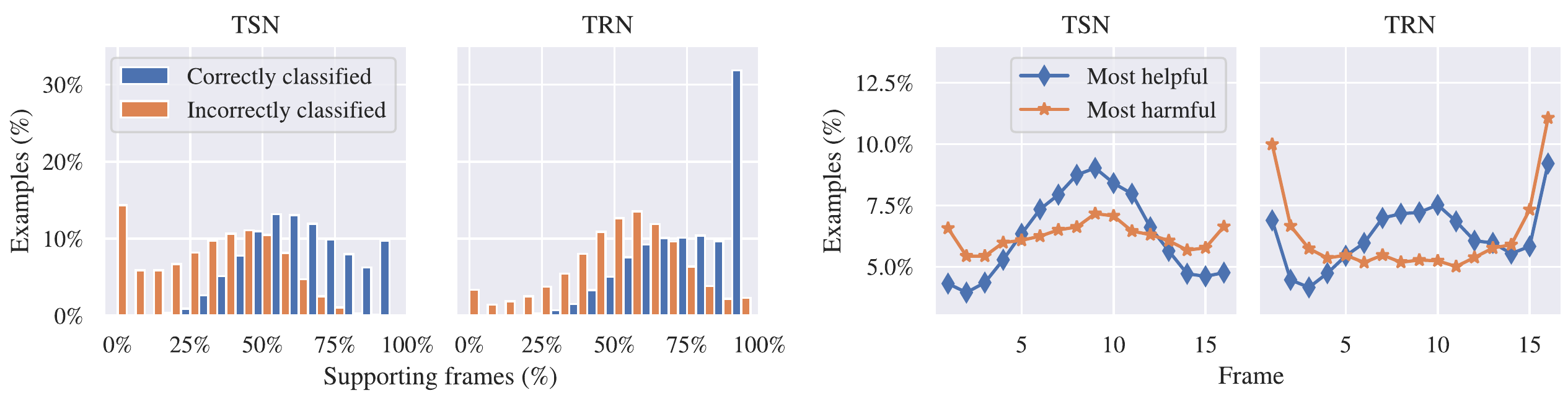}
        \put (0, 2) {(a)}
        \put (53, 2) {(b)}
      \end{overpic}
      \phantomsubcaption
    \label{fig:shapley-value:positive-contribution-distribution}
    \phantomsubcaption
    \label{fig:helpful-harmful-frames-by-position}
  \end{minipage}
  \vspace{-2em}
  \caption{
    \subref{fig:shapley-value:positive-contribution-distribution}
    The percentage of supporting frames for correctly/incorrectly classified examples, comparing TSN to TRN.
    \subref{fig:helpful-harmful-frames-by-position} \% of examples where the frame is the most supporting/distracting show that TSN/TRN make use of frames in different positions.
  }
\end{figure}

We perform the same test, discarding frames by attribution rank for ESV, IG and GC in \cref{fig:method-comparison-frame-ablation}.
When discarding 4 frames by their decreasing rank, the model's accuracy increases by 20\% for ESV compared to 8\% for GC and 7\% for IG.
These results demonsrate that ESV's attribution values are more representative of how the model values frames compared to the other attribution methods.

\question{What can we learn from \textit{supporting} and \textit{distracting} frames?}
We analyse the proportion of supporting frames across correctly and incorrectly classified examples in \cref{fig:shapley-value:positive-contribution-distribution}, comparing TRN to TSN.
Correctly classified examples are more likely to have a larger proportion of supporting frames, however a number of correctly classified videos contain \textit{distracting} frames.
There are more supporting frames for TRN compared to TSN for correctly classified examples.

\begin{figure}[t]
\vspace{-5pt}
  \centering
  \includegraphics[width=.31\textwidth]{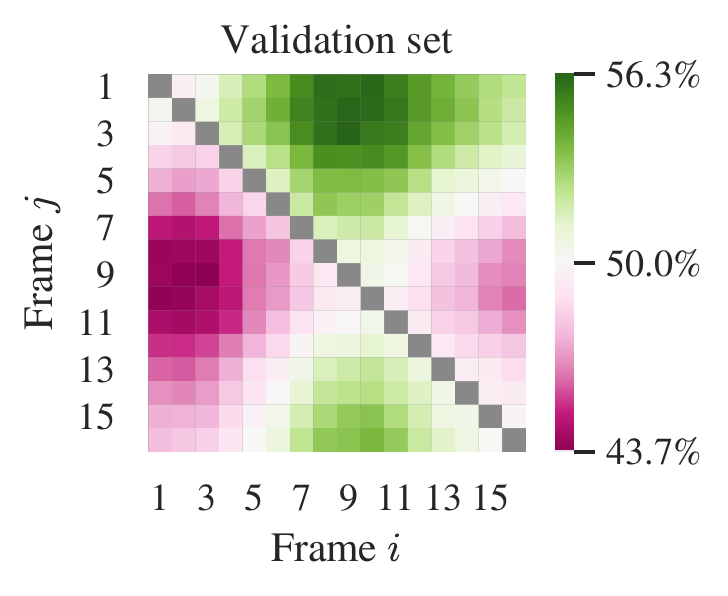}
  \includegraphics[width=.31\textwidth]{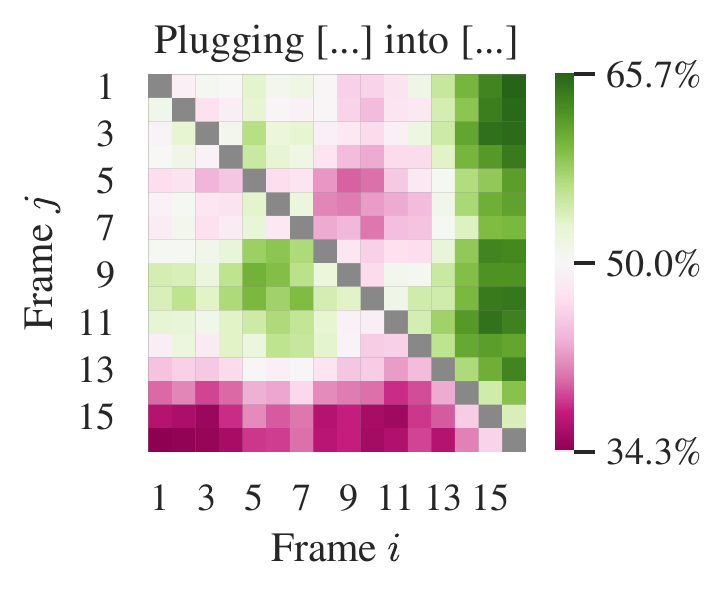}
  \includegraphics[width=.33\textwidth]{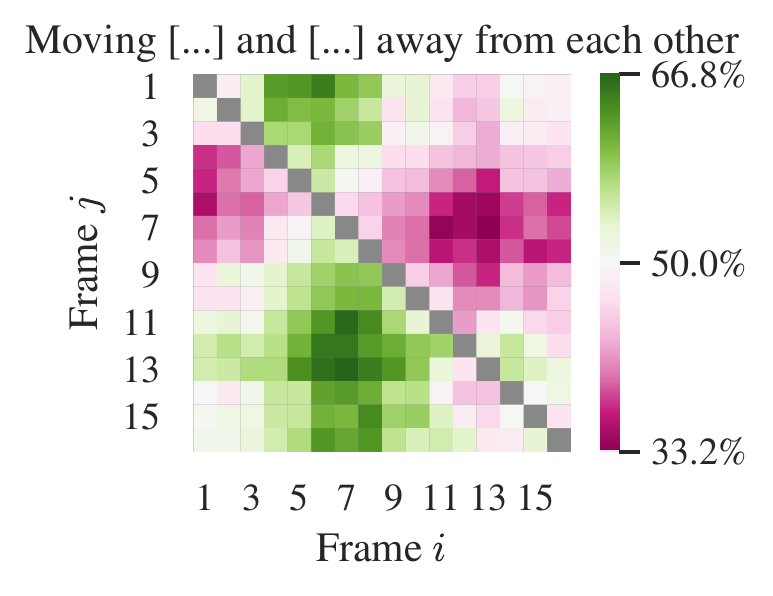}
  \vspace{-1em}
  \caption{
    Comparing the percentage of videos where $\phi^c_\cci > \phi^c_\ccj$ for TRN.
    This shows that, across the dataset, some frames are on average more informative than others, and for certain classes this can be quite different to the overall trend.
    }
  \label{fig:shapley-value:comparing-frames-shapley-values-by-location}
\end{figure}

\question{Is there a correlation between a frame's position in the sequence and its ESV?}
We consider the proportion of videos where each frame, by its position in the sequence, is the most helpful or harmful in \cref{fig:helpful-harmful-frames-by-position}.
The first/last frames are often the most impactful for TRN, as the model learns temporal relationships.

We further analyse TRN, reporting the percentage of videos for which $\phi^c_\cci > \phi^c_\ccj$ for all combinations of frame positions $i$ and $j$ across the validation set as well as per-class.
\cref{fig:shapley-value:comparing-frames-shapley-values-by-location} shows that
(i) frames residing in the middle of the sequence have higher attributions on average than those at the edges;
(ii) class trends can deviate from this distinctly. The later frames for the class ``Plugging [...] into [...]'' are most important, as these are the ones that discriminate it from the similar action ``Plugging [...] into [...] but pulling it right out'';
(iii) Early frames from ``Moving [...] and [...] away from each other'' contribute most, as the objects move further from one another as the video progresses.

\begin{figure}[t]
  \centering
  \begin{minipage}[t]{0.375\textwidth}
    \vspace{0pt}  
    \begin{overpic}[width=\textwidth]{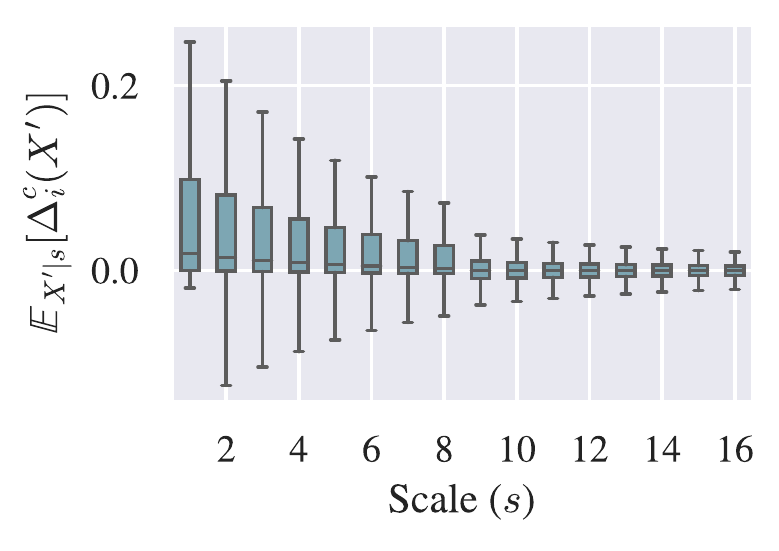}
    \put (2, 8) {(a)}
    \end{overpic}
    \phantomsubcaption
    \label{fig:average-marginal-contributions-across-scales:trn}
  \end{minipage}
   \begin{minipage}[t]{0.27\textwidth}
    \vspace{0pt}  
    \begin{overpic}[width=\textwidth]{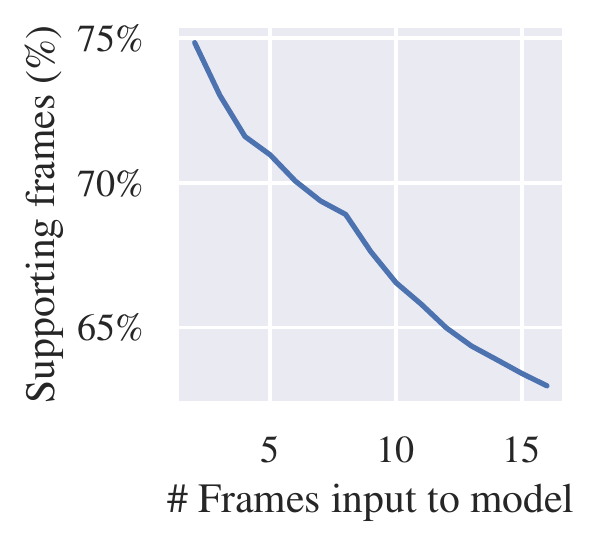}
    \put (2, 3) {(b)}
    \end{overpic}
    \phantomsubcaption
    \label{fig:shapley-value:varying-number-of-input-frames:trn}
  \end{minipage}
  \begin{minipage}[t]{0.265\textwidth}
    \vspace{0pt}  
    \begin{overpic}[width=\textwidth]{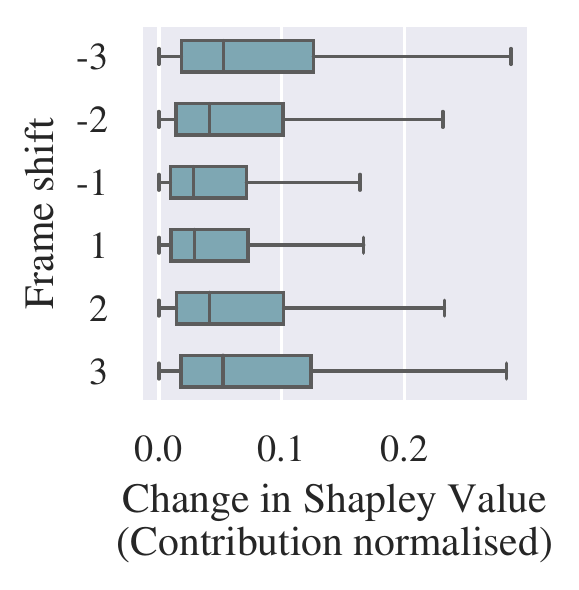}
    \put (2, 10) {(c)}
    \end{overpic}
    \phantomsubcaption
    \label{fig:temporal-stability:trn}
  \end{minipage}
  \vspace{-2.2em}
  \caption{%
    TRN: \subref{fig:average-marginal-contributions-across-scales:trn}
    Box-plot of marginal contributions at each scale.
    \subref{fig:shapley-value:varying-number-of-input-frames:trn} Increasing the number of frames fed to the model decreases the percentage of supporting frames.
    \subref{fig:temporal-stability:trn}~Change in ESV when compared to neighbouring frames.
  }
  \label{fig:scale-analysis:trn}
\end{figure}

\question{How do subsequences of various lengths contribute to the ESV?}
Since ESV is the average of the marginal contributions of subsequences at each scale, we analyse the per-scale average marginal contributions to probe the contribution of subsequences of a certain length $\ccs$ (\cref{fig:average-marginal-contributions-across-scales:trn}).
The average marginal contributions steadily decreases as longer subsequences are considered, indicating that the majority of the frame's attribution is already extracted from shorter subsequences.
We then consider whether there is a relationship between the number of frames fed to the model and the proportion of supporting frames in \cref{fig:shapley-value:varying-number-of-input-frames:trn}.
The longer the subsequence, the more likely it is some frames become redundant and potentially distracting.
This is an interesting finding---it showcases that utilising all frames in the sequence could harm model performance in the presence of distracting frames.

\question{Are ESVs temporally smooth?}
We modify 16-frame sequences by replacing one frame in the uniformly sampled sequence with the frame 1, 2, or 3 places to the left/right. We compute the normalised difference in ESV\footnote{As the contributions vary between sequences, we first normalise the ESVs so that the mass of contributions sum to one before comparing them, \ie{} $\sum |\phi^c_i| = 1$.} for the original and replaced frame.
We plot the distribution of changes in \cref{fig:temporal-stability:trn}.
The figure shows a general trend of temporal smoothness, symmetric for both left/right shifts; the further the shift in either direction, the greater the change in ESV.

\begin{figure}[t]
  \centering
  \includegraphics[width=\textwidth]{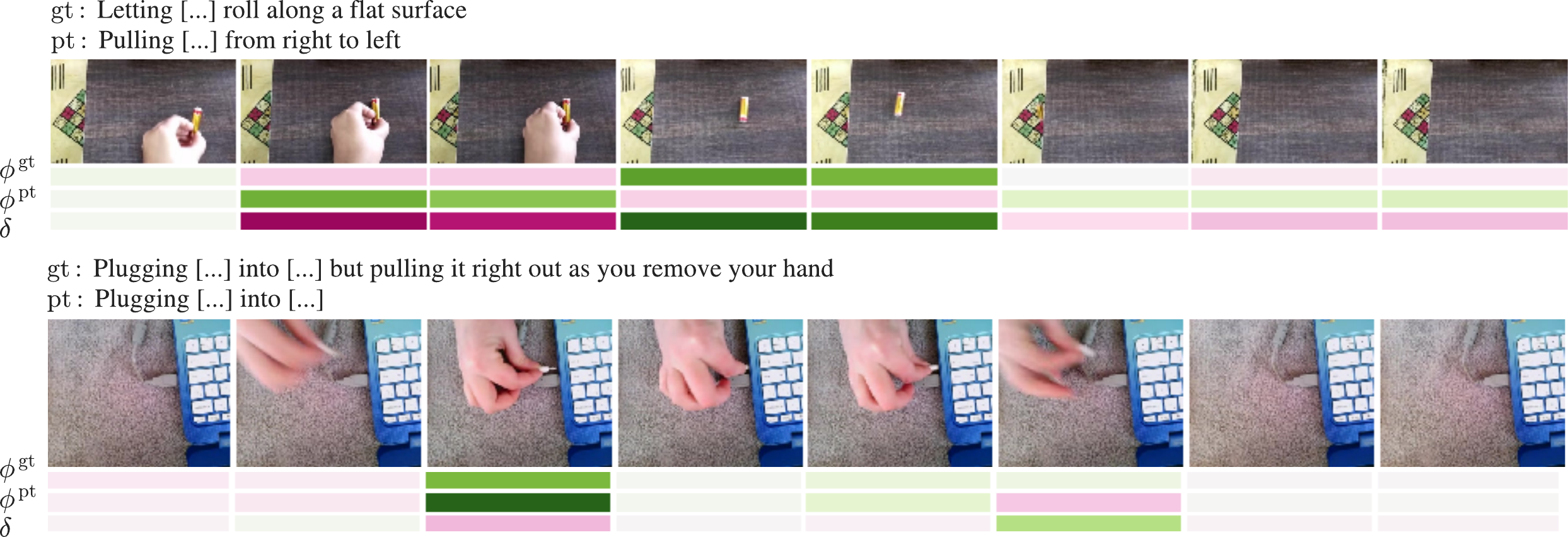}
  \vspace{-1.5em}
  \caption{
      Class-contrastive ESVs comparing the ground truth class (gt) to the predicted class (pt).
      \protect\includegraphics[height=.7em]{./media/cbar.pdf}
  }
  \label{fig:contrastive-shapley-values}

\end{figure}

\question{What about class-contrastive ESVs?}
Up to this point we have focused on ESVs for one class, the ground truth class of a video.
We present qualitative examples of class-contrastive ESVs in \cref{fig:contrastive-shapley-values}.
These show which frames contribute more to the ground-truth class than the predicted class.
In the top sequence, the first frames confuse the model contributing more to ``Pulling [...] from left to right''. Frames when the battery is rolling contribute highly to the ground truth class.
The second sequence shows that plugging the cable into the laptop contributes to both the predicted and ground truth class, but there is insufficient support from the frame where the cable is unplugged to make the correct classification. 

\begin{figure}[t]
  \vspace{-.5em}
  \centering
  \includegraphics[width=\textwidth]{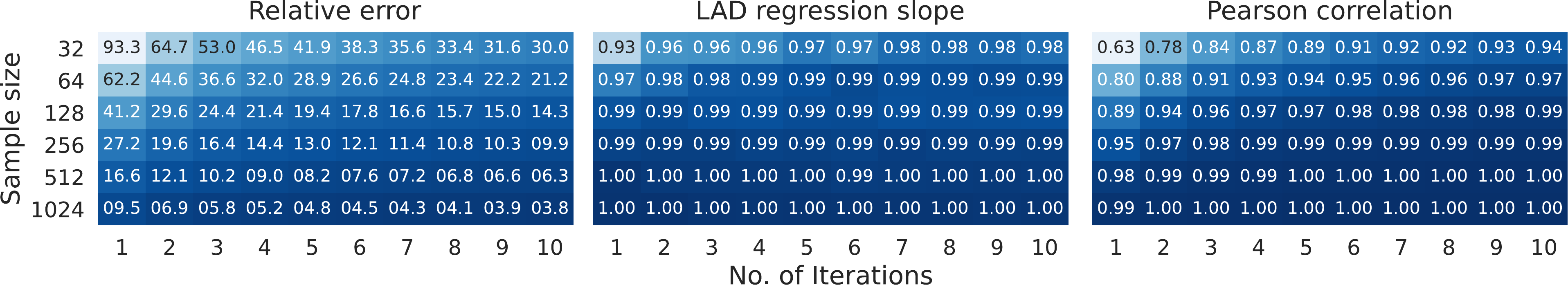}
  \begin{center}
  \vspace{-5pt}
    \begin{tabular}{l@{\hspace{12pt}} *{7}{@{\hspace{6pt}}r}}
      \toprule
      Scale sample size $m$                    & 32   & 64   & 128  & 256  & 512  & 1024\\
      \midrule
      \% of total subsequences/iteration  & 0.68 & 1.32 & 2.56 & 4.71 & 9.01 & 16.19\\
      \bottomrule
    \end{tabular}
  \end{center}
  \vspace{-1.2em}
  \caption{
    Comparing approximate vs.\ exact ESVs for 16-frame videos, as we vary $m$, the maximum number of subsequences at each scale, (y-axis) and the number of iterations~(x-axis).
    For relative error, lower is better, and for LAD regression slope and Pearson correlation, closer to one is better.
    The table shows \% of subsequences sampled per iteration as $m$ varies.
  }
  \label{fig:approximation-evaluation-heatmaps}
\end{figure}

\subsection{ESV Approximation Evaluation}
\label{sec:experiment:approximation-evaluation}

We now evaluate our ability to approximate Shapley values through our tractable approach proposed in \cref{sec:approximation}.
We first compare approximate to exact ESVs on 16 frame sequences (longer sequences limited by GPU memory).
We then scale up our analysis to compute ESVs for all frames in variable-length videos and show approximate ESVs are consistent with those computed exactly for shorter sequences.

\begin{figure}[t]
    \centering
    \includegraphics[width=\textwidth]{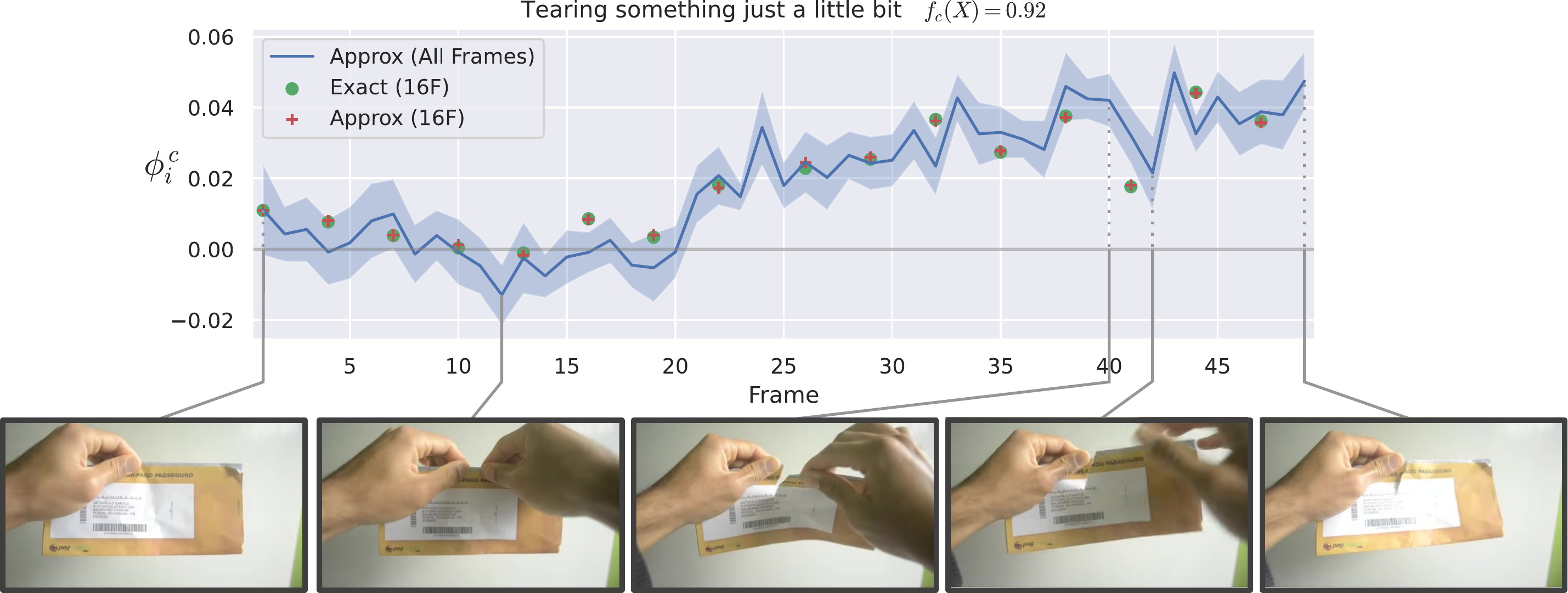}
    \caption{
    ESVs for all frames computed using our approximation method (we plot the mean ESVs over 10 runs of our algorithm and shade  $\mu\pm\sigma$, $m = 1024$ with 4 iterations).
    We also plot the approximate and exact ESVs for 16 uniformly sampled frames demonstrating these are representative of the ESVs for longer sequences.
    }
    \label{fig:all-frames-approximation-qualitative}
\end{figure}

In \cref{alg:shapley-value-computation}, we sample $m$ subsequences per scale every iteration.
We compare the approximate ESV $\hat{\phi}^c_i$ produced by subsequence sampling against exact ESV $\phi^c_i$, to assess the error introduced by approximation.
For this assessment, we randomly sample 1,000 videos where $f'_c(X) \geq 0.05$ so the ESVs aren't so small as to compromise the approximation evaluation. We then uniformly sample 16 frames from each video.
We consider three evaluation metrics to assess the effect of sampling:
%
\begin{enumerate}
\item \textit{Relative error}, we compute the normalised error between approximate and exact ESVs per element $|(\hat{\phi}^c_\cci - \phi^c_\cci)| / A$ where $A = \frac{1}{|X|}\sum_\ccj|\phi^c_\ccj|$ is the video-level mean of the unsigned ESVs.
\item \textit{Bias}, we fit a Least Absolute Deviance (LAD) regression between $\phi^c_\cci$ and $\hat\phi^c_\cci$. A regression slope $< 1$ shows an over-estimate.
\item \textit{Correlation}, we compute Pearson's $r$ between the approximate and exact ESVs computed for each video.
\end{enumerate}
These metrics are computed per video and averaged across all videos.
We present the results in \cref{fig:approximation-evaluation-heatmaps} and demonstrate the efficiency of the sampling by reporting the percentage of subsequences sampled per iteration, for various $m$ values in the table underneath.
For instance, at $m=256$ over 4 iterations, we would have considered less than 19\% of all subsequences, but would achieve $r = 0.99$ and no bias in the approximation.
Increasing the sample size $m$ and/or number of iterations improves all metrics.
Similar to \cref{fig:concept}, in \cref{fig:all-frames-approximation-qualitative} we plot ESVs computed for all 49 frames in a video.
Without approximation we would require $10^{14}$ evaluations of $\ccfms$, but with our approximation requires only 49k evaluations.

\section{Related Work}
\label{sec:related-work}

\paragraph{Feature Attribution.}
These techniques can be categorised into backprop~\cite{selvaraju2017_GradCAMVisualExplanations,chattopadhay2018_GradCAMGeneralizedGradientBased,bach2015_PixelWiseExplanationsNonLinear,adelbargal2018_ExcitationBackpropRNNs,sundararajan2017_AxiomaticAttributionDeep,simonyan2014_DeepConvolutionalNetworks,zeiler2014_VisualizingUnderstandingConvolutional,bach2015_PixelWiseExplanationsNonLinear,shrikumar2019_LearningImportantFeatures} or perturbation based~\cite{zeiler2014_VisualizingUnderstandingConvolutional,lundberg2017_UnifiedApproachInterpreting,fong2017_InterpretableExplanationsBlack,petsiuk2018_RISERandomizedinputa,fong2019_UnderstandingDeepNetworks} methods.
Backprop methods
either use standard or modified backpropagation rules to work out a feature's contribution to an output neuron.
These are more attractive computationally, as they don't require multiple model evaluations, but often don't satisfy expected principles of an attribution method such as implementation invariance, sensitivity to the input or the model parameters~\cite{adebayo2018_SanityChecksSaliency,sundararajan2017_AxiomaticAttributionDeep}.
Grad-CAM\cite{selvaraju2017_GradCAMVisualExplanations} is most popular, but a variety of others exist~\cite{chattopadhay2018_GradCAMGeneralizedGradientBased,bach2015_PixelWiseExplanationsNonLinear,shrikumar2019_LearningImportantFeatures,zhang2018_TopdownNeuralAttention,simonyan2014_DeepConvolutionalNetworks,zeiler2014_VisualizingUnderstandingConvolutional}.

In contrast, perturbation-based methods evaluate the model on perturbed inputs.
Occlusion~\cite{zeiler2014_VisualizingUnderstandingConvolutional} was first utilised, sliding an occluding patch over the image and evaluating the change in the model's output.
\u{S}trumbelj and Kononenko~\cite{strumbelj2010_EfficientExplanationIndividual}  proposed the use of Shapley values as a model explanation, initially retraining a model on every possible subset of features, but later~\cite{strumbelj2014_Explainingpredictionmodels} using feature substitution.
SHAP~\cite{lundberg2017_UnifiedApproachInterpreting} unified various attribution methods~\cite{ribeiro2016_WhyShouldTrust,shrikumar2019_LearningImportantFeatures,bach2015_PixelWiseExplanationsNonLinear} by showing how each could compute or approximate Shapley values under certain assumptions (\eg{} feature independence, model linearity).

Recent hybrid approaches combine perturbation with backprop~\cite{sundararajan2017_AxiomaticAttributionDeep,fong2017_InterpretableExplanationsBlack,fong2019_UnderstandingDeepNetworks}.
Integrated gradients (IG)~\cite{sundararajan2017_AxiomaticAttributionDeep} uses Aumann-Shapley values~\cite{aumann1974_ValuesNonatomicGames} to compute feature attributions for differentiable models, however the method requires a reference and evaluates the model on out-of-distribution examples generated by linearly interpolating between the input and the reference.
Meaningful perturbations~\cite{fong2017_InterpretableExplanationsBlack} learn a mask to retain the most important pixels in an input image to a class neuron. They add regularisers to encourage sparsity and contiguous region formation.
Extremal perturbations~\cite{fong2019_UnderstandingDeepNetworks} address some of the shortcomings of~\cite{fong2017_InterpretableExplanationsBlack}  such as the non-uniqueness of masks.
Both works~\cite{fong2017_InterpretableExplanationsBlack,fong2019_UnderstandingDeepNetworks} produce binary masks, rather than pixel attributions.

\paragraph{Explaining Video Models.}

Most works~\cite{chattopadhay2018_GradCAMGeneralizedGradientBased,doughty2018_WhoBetterWho,goyal2018_Evaluatingvisualcommona,adelbargal2018_ExcitationBackpropRNNs,srinivasan2017_Interpretablehumanaction,manttari2020_Interpretingvideofeatures} in attribution for video understanding use backprop methods originally designed for images, such as Grad-CAM~\cite{selvaraju2017_GradCAMVisualExplanations} or EBP~\cite{zhang2018_TopdownNeuralAttention}.
A few recent works propose video-specific attribution methods.
EBP-RNN\cite{adelbargal2018_ExcitationBackpropRNNs} extends EBP~\cite{zhang2018_TopdownNeuralAttention} to CNN+RNN to explain video models.
Saliency tubes\cite{stergiou2019_SaliencyTubesVisual} use a Grad-CAM like formulation to provide spatio-temporal attributions of the last convolutional layer. 
They extend the approach in~\cite{stergiou2019_ClassFeaturePyramids} by backpropagating gradients to different depths in the network.
M\"antt\"ari \etal{}~\cite{manttari2020_Interpretingvideofeatures} apply meaningful perturbations~\cite{fong2017_InterpretableExplanationsBlack} to learn a temporal mask over the input.
To keep the number of input frames fixed, they replace missing frames by duplication.
Li \etal{}~\cite{li2020_ComprehensiveStudyVisual} learn a spatio-temporal mask via extremal perturbations~\cite{fong2019_UnderstandingDeepNetworks}.
They replace missing voxels with a reference of a blurred voxel,
analysing R(2+1)D~\cite{xie2018_RethinkingSpatiotemporalFeature} and a CNN+LSTM model.

In contrast to these works,
we offer the first perturbation method for video explanation that is based on the Shapley axioms of fairness. Our method obeys principled criteria (\cref{sec:element-attribution:feature-attribution}) and does not feed models with out-of-distribution examples as in~\cite{sundararajan2017_AxiomaticAttributionDeep,fong2017_InterpretableExplanationsBlack,fong2019_UnderstandingDeepNetworks}.

Alternate efforts in explaining video models take a network-centric approach.
Feichtenhofer \etal{}~\cite{feichtenhofer2020_DeepInsightsConvolutional} use activation maximisation~\cite{simonyan2014_DeepConvolutionalNetworks} to explain a variety of two-stream architectures~\cite{feichtenhofer2016_SpatiotemporalResidualNetworks,feichtenhofer2016_ConvolutionalTwoStreamNetwork,wang2016_TemporalSegmentNetworks}, synthesizing inputs that maximally activate a chosen class neuron.
Huang \etal{}~\cite{huang2018_WhatMakesVideo} explain the temporal signal in video by training a temporal generative model to produce within-distribution videos of shorter or re-ordered frames. They explain C3D~\cite{tran2015_LearningSpatiotemporalFeatures} on a fixed-length sequence, analysing the model's performance drop.
These approaches offer a complimentary viewpoint to model explanations, and do not attempt frame attributions.

\section{Conclusion}
\label{sec:conclusion}

In this paper, we introduced the notion of element attribution for determining the contribution of each frame to the output of an action recognition model.
We adopted the Shapley axioms as a way of determining these contributions in a fair and principled manner in our proposed Element Shapley Value (ESV).
We side-step the issues present in feature-attribution by reformulating the Shapley value and utilise multi-scale models to determine the marginal contributions of frames without having to substitute them.
We used ESV to analyse frame-based action recognition models on the Something-something dataset, showing insights into how trained models make classification decisions.

\vspace{8pt}
\noindent \textbf{Acknowledgement.} Research supported by EPSRC UMPIRE (EP/T004991/1) and EPSRC Doctoral Training Partnershipts (DTP).

\clearpage

\bibliographystyle{splncs}
\bibliography{bibliography}

\begin{thebibliography}{10}

\bibitem{zhou2018_TemporalRelationalReasoning}
Zhou, B., Andonian, A., Oliva, A., Torralba, A.:
\newblock Temporal {{Relational Reasoning}} in {{Videos}}.
\newblock In: Proceedings of the {{European Conference}} on {{Computer Vision}}
  ({{ECCV}}). (2018)  803--818

\bibitem{tran2018_CloserLookSpatiotemporal}
Tran, D., Wang, H., Torresani, L., Ray, J., LeCun, Y., Paluri, M.:
\newblock A {{Closer Look}} at {{Spatiotemporal Convolutions}} for {{Action
  Recognition}}.
\newblock In: Proceedings of the {{IEEE Conference}} on {{Computer Vision}} and
  {{Pattern Recognition}} ({{CVPR}}). (2018)  6450--6459

\bibitem{hussein2019_TimeceptionComplexAction}
Hussein, N., Gavves, E., Smeulders, A.W.M.:
\newblock Timeception for {{Complex Action Recognition}}.
\newblock In: Proceedings of the {{IEEE Conference}} on {{Computer Vision}} and
  {{Pattern Recognition}} ({{CVPR}}). (2019)  254--263

\bibitem{girdhar2019_VideoActionTransformer}
Girdhar, R., Carreira, J., Doersch, C., Zisserman, A.:
\newblock Video {{Action Transformer Network}}.
\newblock In: Proceedings of the {{IEEE Conference}} on {{Computer Vision}} and
  {{Pattern Recognition}} ({{CVPR}}). (2019)  244--253

\bibitem{chen2018_MultifiberNetworksVideo}
Chen, Y., Kalantidis, Y., Li, J., Yan, S., Feng, J.:
\newblock Multi-fiber {{Networks}} for {{Video Recognition}}.
\newblock In: Proceedings of the {{European Conference}} on {{Computer Vision}}
  ({{ECCV}}). (2018)

\bibitem{lin2019_TSMTemporalShift}
Lin, J., Gan, C., Han, S.:
\newblock {{TSM}}: {{Temporal Shift Module}} for {{Efficient Video
  Understanding}}.
\newblock In: Proceedings of the {{IEEE Conference}} on {{Computer Vision}} and
  {{Pattern Recognition}} ({{CVPR}}). (2019)  7083--7093

\bibitem{feichtenhofer2019_SlowFastNetworksVideo}
Feichtenhofer, C., Fan, H., Malik, J., He, K.:
\newblock {{SlowFast Networks}} for {{Video Recognition}}.
\newblock In: Proceedings of the {{IEEE Conference}} on {{Computer Vision}} and
  {{Pattern Recognition}} ({{CVPR}}). (2019)  6202--6211

\bibitem{abu-el-haija2016_YouTube8MLargeScaleVideo}
{Abu-El-Haija}, S., Kothari, N., Lee, J., Natsev, P., Toderici, G.,
  Varadarajan, B., Vijayanarasimhan, S.:
\newblock {{YouTube}}-{{8M}}: {{A Large}}-{{Scale Video Classification
  Benchmark}}.
\newblock ArXiv160908675 Cs (2016)

\bibitem{kay2017_KineticsHumanAction}
Kay, W., Carreira, J., Simonyan, K., Zhang, B., Hillier, C., Vijayanarasimhan,
  S., Viola, F., Green, T., Back, T., Natsev, P., Suleyman, M., Zisserman, A.:
\newblock The {{Kinetics Human Action Video Dataset}}.
\newblock ArXiv170506950 Cs (2017)

\bibitem{goyal2017_SomethingSomethingVideo}
Goyal, R., Ebrahimi~Kahou, S., Michalski, V., Materzynska, J., Westphal, S.,
  Kim, H., Haenel, V., Fruend, I., Yianilos, P., {Mueller-Freitag}, M., Hoppe,
  F., Thurau, C., Bax, I., Memisevic, R.:
\newblock The ``{{Something Something}}'' {{Video Database}} for {{Learning}}
  and {{Evaluating Visual Common Sense}}.
\newblock In: Proceedings of the {{IEEE International Conference}} on
  {{Computer Vision}} ({{ICCV}}). (2017)

\bibitem{diba2019_LargeScaleHolistic}
Diba, A., Fayyaz, M., Sharma, V., Paluri, M., Gall, J., Stiefelhagen, R.,
  Van~Gool, L.:
\newblock Large {{Scale Holistic Video Understanding}}.
\newblock In: Proceedings of the {{European Conference}} on {{Computer Vision}}
  ({{ECCV}}). (2020)

\bibitem{gu2018_AVAVideoDataset}
Gu, C., Sun, C., Ross, D.A., Vondrick, C., Pantofaru, C., Li, Y.,
  Vijayanarasimhan, S., Toderici, G., Ricco, S., Sukthankar, R., Schmid, C.,
  Malik, J.:
\newblock {{AVA}}: {{A Video Dataset}} of {{Spatio}}-{{Temporally Localized
  Atomic Visual Actions}}.
\newblock In: Proceedings of the {{IEEE Conference}} on {{Computer Vision}} and
  {{Pattern Recognition}} ({{CVPR}}). (2018)  6047--6056

\bibitem{damen2018_ScalingEgocentricVisionb}
Damen, D., Doughty, H., Maria~Farinella, G., Fidler, S., Furnari, A., Kazakos,
  E., Moltisanti, D., Munro, J., Perrett, T., Price, W., Wray, M.:
\newblock Scaling {{Egocentric Vision}}: {{The EPIC}}-{{KITCHENS Dataset}}.
\newblock In: Proceedings of the {{European Conference}} on {{Computer Vision}}
  ({{ECCV}}). (2018)  720--736

\bibitem{strumbelj2010_EfficientExplanationIndividual}
{\v S}trumbelj, E., Kononenko, I.:
\newblock An {{Efficient Explanation}} of {{Individual Classifications}} using
  {{Game Theory}}.
\newblock J. Mach. Learn. Res. \textbf{11} (2010)  1--18

\bibitem{strumbelj2014_Explainingpredictionmodels}
{\v S}trumbelj, E., Kononenko, I.:
\newblock Explaining prediction models and individual predictions with feature
  contributions.
\newblock Knowl Inf Syst \textbf{41} (2014)  647--665

\bibitem{lundberg2017_UnifiedApproachInterpreting}
Lundberg, S.M., Lee, S.I.:
\newblock A {{Unified Approach}} to {{Interpreting Model Predictions}}.
\newblock In Guyon, I., Luxburg, U.V., Bengio, S., Wallach, H., Fergus, R.,
  Vishwanathan, S., Garnett, R., eds.: Advances in {{Neural Information
  Processing Systems}} 30 ({{NeurIPS}}).
\newblock {Curran Associates, Inc.} (2017)  4765--4774

\bibitem{sundararajan2017_AxiomaticAttributionDeep}
Sundararajan, M., Taly, A., Yan, Q.:
\newblock Axiomatic {{Attribution}} for {{Deep Networks}}.
\newblock In: Proceedings of the 34th {{International Conference}} on {{Machine
  Learning}} ({{ICML}}). {{ICML}}'17, {Sydney, NSW, Australia}, {JMLR.org}
  (2017)  3319--3328

\bibitem{sundararajan2020_manyShapleyvalues}
Sundararajan, M., Najmi, A.:
\newblock The many {{Shapley}} values for model explanation.
\newblock In: Proceedings of the 37th {International {{Conference}} on
  {{Machine Learning}} ({{ICML}})}. {{ICML}}'20 (2020)

\bibitem{shapley1953_ValuenPersonGames}
Shapley, L.S.:
\newblock A {{Value}} for n-{{Person Games}}.
\newblock In: Contributions to the {{Theory}} of {{Games}} ({{AM}}-28),
  {{Volume II}}. Volume~2.
\newblock {Princeton University Press}, {Princeton} (1953)

\bibitem{selvaraju2017_GradCAMVisualExplanations}
Selvaraju, R.R., Cogswell, M., Das, A., Vedantam, R., Parikh, D., Batra, D.:
\newblock Grad-{{CAM}}: {{Visual Explanations From Deep Networks}} via
  {{Gradient}}-{{Based Localization}}.
\newblock In: Proceedings of the {{IEEE International Conference}} on
  {{Computer Vision}} ({{ICCV}}). (2017)  618--626

\bibitem{zhang2018_TopdownNeuralAttention}
Zhang, J., Bargal, S.A., Lin, Z., Brandt, J., Shen, X., Sclaroff, S.:
\newblock Top-{{Down Neural Attention}} by {{Excitation Backprop}}.
\newblock Int J Comput Vis \textbf{126} (2018)  1084--1102

\bibitem{fong2019_UnderstandingDeepNetworks}
Fong, R., Patrick, M., Vedaldi, A.:
\newblock Understanding {{Deep Networks}} via {{Extremal Perturbations}} and
  {{Smooth Masks}}.
\newblock In: Proceedings of the {{IEEE International Conference}} on
  {{Computer Vision}} ({{ICCV}}). (2019)  2950--2958

\bibitem{bach2015_PixelWiseExplanationsNonLinear}
Bach, S., Binder, A., Montavon, G., Klauschen, F., M{\"u}ller, K.R., Samek, W.:
\newblock On {{Pixel}}-{{Wise Explanations}} for {{Non}}-{{Linear Classifier
  Decisions}} by {{Layer}}-{{Wise Relevance Propagation}}.
\newblock PLOS ONE \textbf{10} (2015)  e0130140

\bibitem{fong2017_InterpretableExplanationsBlack}
Fong, R.C., Vedaldi, A.:
\newblock Interpretable {{Explanations}} of {{Black Boxes}} by {{Meaningful
  Perturbation}}.
\newblock In: Proceedings of the {{IEEE International Conference}} on
  {{Computer Vision}} ({{ICCV}}). (2017)  3429--3437

\bibitem{ribeiro2016_WhyShouldTrust}
Ribeiro, M.T., Singh, S., Guestrin, C.:
\newblock "{{Why Should I Trust You}}?": {{Explaining}} the {{Predictions}} of
  {{Any Classifier}}.
\newblock In: Proceedings of the 22nd {{ACM International Conference}} on
  {{Knowledge Discovery}} and {{Data Mining}} ({{SIGKDD}}). {{KDD}} '16, {San
  Francisco, California, USA}, {Association for Computing Machinery} (2016)
  1135--1144

\bibitem{carreira2017_QuoVadisAction}
Carreira, J., Zisserman, A.:
\newblock Quo {{Vadis}}, {{Action Recognition}}? {{A New Model}} and the
  {{Kinetics Dataset}}.
\newblock In: Proceedings of the {{IEEE Conference}} on {{Computer Vision}} and
  {{Pattern Recognition}} ({{CVPR}}). (2017)  6299--6308

\bibitem{young1985_Monotonicsolutionscooperative}
Young, H.P.:
\newblock Monotonic solutions of cooperative games.
\newblock Int J Game Theory \textbf{14} (1985)  65--72

\bibitem{strumbelj2009_Explaininginstanceclassifications}
{\v S}trumbelj, E., Kononenko, I., Robnik~{\v S}ikonja, M.:
\newblock Explaining instance classifications with interactions of subsets of
  feature values.
\newblock Data Knowl. Eng. \textbf{68} (2009)  886--904

\bibitem{manttari2020_Interpretingvideofeatures}
M{\"a}ntt{\"a}ri, J., Broom{\'e}, S., Folkesson, J., Kjellstr{\"o}m, H.:
\newblock Interpreting video features: A comparison of {{3D}} convolutional
  networks and convolutional {{LSTM}} networks.
\newblock ArXiv200200367 Cs (2020)

\bibitem{li2020_ComprehensiveStudyVisual}
Li, Z., Wang, W., Li, Z., Huang, Y., Sato, Y.:
\newblock A {{Comprehensive Study}} on {{Visual Explanations}} for
  {{Spatio}}-temporal {{Networks}}.
\newblock ArXiv200500375 Cs (2020)

\bibitem{sturmfels2020_VisualizingImpactFeature}
Sturmfels, P., Lundberg, S., Lee, S.I.:
\newblock Visualizing the {{Impact}} of {{Feature Attribution Baselines}}.
\newblock Distill \textbf{5} (2020)  e22

\bibitem{huang2018_WhatMakesVideo}
Huang, D.A., Ramanathan, V., Mahajan, D., Torresani, L., Paluri, M., {Fei-Fei},
  L., Carlos~Niebles, J.:
\newblock What {{Makes}} a {{Video}} a {{Video}}: {{Analyzing Temporal
  Information}} in {{Video Understanding Models}} and {{Datasets}}.
\newblock In: Proceedings of the {{IEEE Conference}} on {{Computer Vision}} and
  {{Pattern Recognition}} ({{CVPR}}). (2018)  7366--7375

\bibitem{dwibedi2018_TemporalReasoningVideos}
Dwibedi, D., Sermanet, P., Tompson, J.:
\newblock Temporal {{Reasoning}} in {{Videos Using Convolutional Gated
  Recurrent Units}}.
\newblock In: Proceedings of the {{IEEE Conference}} on {{Computer Vision}} and
  {{Pattern Recognition Workshops}} ({{CVPRW}}). (2018)  1111--1116

\bibitem{wang2016_TemporalSegmentNetworks}
Wang, L., Xiong, Y., Wang, Z., Qiao, Y., Lin, D., Tang, X., Van~Gool, L.:
\newblock Temporal segment networks: {{Towards}} good practices for deep action
  recognition.
\newblock In: Proceedings of the {{European Conference}} on {{Computer Vision}}
  ({{ECCV}}), {Springer} (2016)  20--36

\bibitem{chattopadhay2018_GradCAMGeneralizedGradientBased}
Chattopadhay, A., Sarkar, A., Howlader, P., Balasubramanian, V.N.:
\newblock Grad-{{CAM}}++: {{Generalized Gradient}}-{{Based Visual
  Explanations}} for {{Deep Convolutional Networks}}.
\newblock In: 2018 {{IEEE Winter Conference}} on {{Applications}} of {{Computer
  Vision}} ({{WACV}}). (2018)  839--847

\bibitem{doughty2018_WhoBetterWho}
Doughty, H., Damen, D., {Mayol-Cuevas}, W.:
\newblock Who's {{Better}}? {{Who}}'s {{Best}}? {{Pairwise Deep Ranking}} for
  {{Skill Determination}}.
\newblock In: Proceedings of the {{IEEE Conference}} on {{Computer Vision}} and
  {{Pattern Recognition}} ({{CVPR}}). (2018)  6057--6066

\bibitem{goyal2018_Evaluatingvisualcommona}
Goyal, R., Mahdisoltani, F., Berger, G., Gharbieh, W., Bax, I., Memisevic, R.:
\newblock Evaluating visual "common sense" using fine-grained classification
  and captioning tasks.
\newblock In: 6th {{International Conference}} on {{Learning Representations}},
  {{Workshop Track Proceedings}} ({{ICLRW}}). (2018)

\bibitem{srinivasan2017_Interpretablehumanaction}
Srinivasan, V., Lapuschkin, S., Hellge, C., M{\"u}ller, K.R., Samek, W.:
\newblock Interpretable human action recognition in compressed domain.
\newblock In: 2017 {{IEEE International Conference}} on {{Acoustics}},
  {{Speech}} and {{Signal Processing}} ({{ICASSP}}). (2017)  1692--1696

\bibitem{aumann1974_ValuesNonatomicGames}
Aumann, R.J., Shapley, L.S.:
\newblock Values of {{Non}}-Atomic {{Games}}.
\newblock {Princeton University Press} (1974)

\bibitem{captum2019github}
Kokhlikyan, N., Miglani, V., Martin, M., Wang, E., Reynolds, J., Melnikov, A.,
  Lunova, N., {Reblitz-Richardson}, O.:
\newblock {{PyTorch}} captum.
\newblock GitHub Repos. (2019)

\bibitem{adelbargal2018_ExcitationBackpropRNNs}
Adel~Bargal, S., Zunino, A., Kim, D., Zhang, J., Murino, V., Sclaroff, S.:
\newblock Excitation {{Backprop}} for {{RNNs}}.
\newblock In: Proceedings of the {{IEEE Conference}} on {{Computer Vision}} and
  {{Pattern Recognition}} ({{CVPR}}). (2018)  1440--1449

\bibitem{simonyan2014_DeepConvolutionalNetworks}
Simonyan, K., Vedaldi, A., Zisserman, A.:
\newblock Deep {{Inside Convolutional Networks}}: {{Visualising Image
  Classification Models}} and {{Saliency Maps}}.
\newblock In Bengio, Y., LeCun, Y., eds.: 2nd {{International Conference}} on
  {{Learning Representations}} ({{ICLR}}). (2014)

\bibitem{zeiler2014_VisualizingUnderstandingConvolutional}
Zeiler, M.D., Fergus, R.:
\newblock Visualizing and {{Understanding Convolutional Networks}}.
\newblock In Fleet, D., Pajdla, T., Schiele, B., Tuytelaars, T., eds.:
  Proceedings of the {{European Conference}} on {{Computer Vision}} ({{ECCV}}).
  Lecture {{Notes}} in {{Computer Science}}, {Cham}, {Springer International
  Publishing} (2014)  818--833

\bibitem{shrikumar2019_LearningImportantFeatures}
Shrikumar, A., Greenside, P., Kundaje, A.:
\newblock Learning {{Important Features Through Propagating Activation
  Differences}}.
\newblock ArXiv170402685 Cs (2019)

\bibitem{petsiuk2018_RISERandomizedinputa}
Petsiuk, V., Das, A., Saenko, K.:
\newblock {{RISE}}: {{Randomized Input Sampling}} for {{Explanation}} of
  {{Black}}-box {{Models}}.
\newblock In: British {{Machine Vision Conference}} 2018 ({{BMVC}}), {BMVA
  Press} (2018)  151

\bibitem{adebayo2018_SanityChecksSaliency}
Adebayo, J., Gilmer, J., Muelly, M., Goodfellow, I., Hardt, M., Kim, B.:
\newblock Sanity {{Checks}} for {{Saliency Maps}}.
\newblock In Bengio, S., Wallach, H., Larochelle, H., Grauman, K.,
  {Cesa-Bianchi}, N., Garnett, R., eds.: Advances in {{Neural Information
  Processing Systems}} 31 ({{NeurIPS}}).
\newblock {Curran Associates, Inc.} (2018)  9505--9515

\bibitem{stergiou2019_SaliencyTubesVisual}
Stergiou, A., Kapidis, G., Kalliatakis, G., Chrysoulas, C., Veltkamp, R.,
  Poppe, R.:
\newblock Saliency {{Tubes}}: {{Visual Explanations}} for {{Spatio}}-{{Temporal
  Convolutions}}.
\newblock In: 2019 {{IEEE International Conference}} on {{Image Processing}}
  ({{ICIP}}). (2019)  1830--1834

\bibitem{stergiou2019_ClassFeaturePyramids}
Stergiou, A., Kapidis, G., Kalliatakis, G., Chrysoulas, C., Poppe, R.,
  Veltkamp, R.:
\newblock Class {{Feature Pyramids}} for {{Video Explanation}}.
\newblock In: Proceedings of the {{IEEE}}/{{CVF International Conference}} on
  {{Computer Vision Workshops}} ({{ICCVW}}). (2019)  4255--4264

\bibitem{xie2018_RethinkingSpatiotemporalFeature}
Xie, S., Sun, C., Huang, J., Tu, Z., Murphy, K.:
\newblock Rethinking {{Spatiotemporal Feature Learning}}: {{Speed}}-{{Accuracy
  Trade}}-offs in {{Video Classification}}.
\newblock In: Proceedings of the {{European Conference}} on {{Computer Vision}}
  ({{ECCV}}). (2018)  305--321

\bibitem{feichtenhofer2020_DeepInsightsConvolutional}
Feichtenhofer, C., Pinz, A., Wildes, R.P., Zisserman, A.:
\newblock Deep {{Insights}} into {{Convolutional Networks}} for {{Video
  Recognition}}.
\newblock Int J Comput Vis \textbf{128} (2020)  420--437

\bibitem{feichtenhofer2016_SpatiotemporalResidualNetworks}
Feichtenhofer, C., Pinz, A., Wildes, R.:
\newblock Spatiotemporal {{Residual Networks}} for {{Video Action
  Recognition}}.
\newblock In Lee, D.D., Sugiyama, M., Luxburg, U.V., Guyon, I., Garnett, R.,
  eds.: Advances in {{Neural Information Processing Systems}} 29 ({{NeurIPS}}).
\newblock {Curran Associates, Inc.} (2016)  3468--3476

\bibitem{feichtenhofer2016_ConvolutionalTwoStreamNetwork}
Feichtenhofer, C., Pinz, A., Zisserman, A.:
\newblock Convolutional {{Two}}-{{Stream Network Fusion}} for {{Video Action
  Recognition}}.
\newblock In: Proceedings of the {{IEEE Conference}} on {{Computer Vision}} and
  {{Pattern Recognition}} ({{CVPR}}). (2016)  1933--1941

\bibitem{tran2015_LearningSpatiotemporalFeatures}
Tran, D., Bourdev, L., Fergus, R., Torresani, L., Paluri, M.:
\newblock Learning {{Spatiotemporal Features}} with {{3D Convolutional
  Networks}}.
\newblock In: Proceedings of the 2015 {{IEEE International Conference}} on
  {{Computer Vision}} ({{ICCV}}). {{ICCV}} '15, {Washington, DC, USA}, {IEEE
  Computer Society} (2015)  4489--4497

\bibitem{ioffe2015_BatchNormalizationAccelerating}
Ioffe, S., Szegedy, C.:
\newblock Batch {{Normalization}}: {{Accelerating Deep Network Training}} by
  {{Reducing Internal Covariate Shift}}.
\newblock In: International {{Conference}} on {{Machine Learning}} ({{ICML}}).
  (2015)  448--456

\bibitem{he2016_DeepResidualLearning}
He, K., Zhang, X., Ren, S., Sun, J.:
\newblock Deep {{Residual Learning}} for {{Image Recognition}}.
\newblock In: Proceedings of the {{IEEE Conference}} on {{Computer Vision}} and
  {{Pattern Recognition}} ({{CVPR}}). (2016)  770--778

\end{thebibliography}

\clearpage

\appendix

\section{Appendix Introduction}

The following sections present supporting material for the main paper.
We first list all our notations for easy referencing in \cref{appendix:notations}.
We then present proofs for the mathematical results in the paper in \cref{appendix:proofs}.
Next, we describe the experimental set-up in
\cref{appendix:experimental-setup} and go on to provide an runtime
analysis in \cref{appendix:computationl-cost} and additional results on
TSN in \cref{appendix:tsn-results}.

\subsection{Notation Reference}
\label{appendix:notations}

{\small{
\begin{tabularx}{\textwidth}{@{}l@{\hskip .1in}l@{\hskip .15in}X@{}}
\toprule
Variable         & Definition                                                     & Description                                                                                                                                                                             \\ \midrule
$X$              & $(x_i)^n_{i=1}$                                                & An ordered sequence comprised of $n$ elements. In our experiments $x_i$ is a frame and $X$ is a video.                                                                                  \\
$x_i$            & $x_i \in X$                                                    & An element of our sequence $X$.                                                                                                                                                         \\
$X'$             & $X' \subseteq X$                                               & A subsequence of $X$, at times this is constrained to be a proper subset ($X' \subset X$). Check surrounding context for constraints.                        \\
$f(X)$           & -                                                              & A model that operates on a sequence $X$ and produces a vector of class scores.                                                                                    \\
$f_c(X)$         & -                                                              & The score for class $c$ produced when the model $f$ is evaluated on the sequence $X$.                                                                                                   \\
$f_c(\emptyset)$ & -                                                              & The score for class $c$ when there is no input, we choose the empirical class distribution over the training set to represent this.                                                     \\
$f_{gt}(X)$      & -                                                              & The ground truth class score.                                                                                                                                       \\
$f_{pt}(X)$      & -                                                              & The predicted class score.                                                                                                                                          \\
$f_{cc}(X)$      & $f_{gt}(X) - f{pt}(X)$                             & The difference between ground truth and predicted class score.                                                                                              \\
$f^s(X)$         & $f^s : \mathbb{R}^{s\times D} \to \mathbb{R}^{C}$              & A single-scale model mapping from a sequence of length $s$ where each element has dimension $D$ to a set of class scores.                                                               \\
$f^\text{ms}(X)$ & $\mathbb{E}_s \left[ \mathbb{E}_{X' | s} [ f^s(X') ] \right]$  & A multi-scale model built from a set of single scale models.                                                                                                                            \\
$\Delta^c_i(X')$ & $f_c(X' \cup \{x_i\}) - f_c(X')$                               & The \textit{marginal contribution} of $x_i$ on the subsequence $X'$ with the condition that $x_i \not\in X'$.                                                          \\
$\phi_i^c$       & $\sum_{X' \subseteq X \setminus \{x_i\}} w(X') \Delta^c_i(X')$ & The Element Shapley Value for $x_i$ with respect to class $c$. \\
$w(X')$          & $\frac{(|X| - |X'| -1)!|X'|!}{|X|!}$                           & The weighting factor used in the Element Shapley Value definition.                                                                                                                      \\
$\hat{\phi}^c_i$ & -                                                              & Approximated Element Shapley Value computed via \cref{alg:shapley-value-computation}.                                                                                  \\
$\delta_i$       & $\phi^{gt}_i - \phi^{pt}_i$                                    & The difference in Element Shapley Value computed w.r.t the ground truth and predicted class. \\
\bottomrule
\end{tabularx}}}

\section{Proofs}
\label{appendix:proofs}

\subsection{Shapley Value Expectation Forms}
\label{appendix:shapley-value-forms}

\subsubsection{Single Expectation Form.}
\label{appendix:shapley-value-forms:single-expectation}
The Shapley value for an element $\ccxi$ from a sequence $\ccsx$ can be interpreted as the expected marginal contribution of $\ccxi$ on a random coalition $\ccsxp \subseteq \ccsx \setminus \{\ccxi\}$, where $\ccsxp$ maintains the order of elements in $\ccsx$.

\begin{proof}
The Shapley value is originally defined~\cite{shapley1953_ValuenPersonGames} as
\begin{equation}
\phi_i = \sum_{\ccsxp \subseteq X \setminus \{\ccxi\}} \frac{(|\ccsx| - |\ccsxp| - 1)! |\ccsxp|!}{|\ccsxp|!} [\ccf(\ccsxp \cup \{\ccxi\}) - \ccf(\ccsx)]\,.
\label{eq:apx:shapley-value}
\end{equation}
We define a random variable $\ccrvxp$ whose probability mass function (pmf) is
\begin{equation}
  p(\ccrvxp) = \frac{(|\ccsx| - |\ccsxp| - 1)! |\ccsxp|!}{|\ccsx|!}\,.
\end{equation}
$p(\ccrvxp)$ is a valid pmf, given all values are non-negative and its sum is equal to one by direct proof.
\begingroup
\allowdisplaybreaks
\begin{align}
  \sum_{\ccrvxp \subseteq X \setminus \{\ccxi\}} p(\ccsxp)
  &= \sum_{\ccrvxp \subseteq X \setminus \{\ccxi\}} \frac{(|\ccsx| - |\ccsxp| - 1)! |\ccsxp|!}{|\ccsx|!} \\
  &= \sum_{\ccs = 0}^{|\ccsx| - 1}  \sum_{\substack{\ccrvxp \subseteq \ccsx \setminus \{\ccxi\}\\|\ccsxp| = \ccs}} \frac{(|\ccsx| - |\ccsxp| - 1)! |\ccsxp|!}{|\ccsx|!} \\
  &= \sum_{\ccs = 0}^{|\ccsx| - 1}   \frac{(|\ccsx| - \ccs - 1)! \ccs!}{|\ccsx|!} \sum_{\substack{\ccrvxp \subseteq \ccsx \setminus \{\ccxi\}\\|\ccsxp| = \ccs}}1 \\
  &= \sum_{\ccs = 0}^{|\ccsx| - 1}  \frac{1}{|\ccsx|}{\binom{|\ccsx| - 1}{\ccs}}^{-1} \sum_{\substack{\ccsxp \subseteq \ccsx \setminus \{\ccxi\}\\|\ccsxp| = \ccs}} 1 \\
&= \frac{1}{|\ccsx|} \sum_{\ccs = 0}^{|\ccsx| - 1}  {\binom{|\ccsx| - 1}{\ccs}}^{-1} \sum_{\substack{\ccsxp \subseteq \ccsx \setminus \{\ccxi\}\\|\ccsxp| = \ccs}} 1 \\
  &= \frac{1}{|\ccsx|} \sum_{\ccs = 0}^{|\ccsx| - 1}  {\binom{|\ccsx| - 1}{\ccs}}^{-1} {\binom{|\ccsx| - 1}{\ccs}} \\
  &= \frac{1}{|\ccsx|} \sum_{\ccs = 0}^{|\ccsx| - 1} 1\\
  &= 1\,.
\end{align}
\endgroup
Consequently, we can see that
\begin{equation}
  \phi_i = \expectation{\ccsxp}{\ccf(\ccsxp \cup \{\ccxi\}) - \ccf(\ccsxp)}
\end{equation}
\end{proof}

\subsubsection{Conditional Expectation Form.}
\label{appendix:shapley-value-forms:conditional-expectation}

The Shapley value can also be formulated as the expectation over the conditional expectation of the marginal contribution of an element on a coalition of size $\ccs$ where all values of $\ccs$ are equally probable.
\begin{equation}
  \expectation{\ccsxp}{\ccf(\ccsxp \cup \{\ccxi\}) - \ccf(\ccsxp)} = \expectation{\ccs}{\expectation{\ccsxp | \ccs}{\ccf(\ccsxp \cup \{\ccxi\}) - \ccf(\ccsxp)}}
  \label{eq:shapley-value:expectation-equality}
\end{equation}
This is a consequence of the law of total expectation, as we split the sample space into $|\ccsx|$ non-empty and non-overlapping partitions, each containing subsets of size $\ccs \in \intrange{0}{|\ccsx| - 1}$.
\begin{proof}
  We show the equivalence by direct proof.
  Since there are $\binom{|\ccsx| - 1}{\ccs}$ instances of $\ccsxp$ of size $\ccs$ the conditional probability of $\ccsxp$ given $\ccs$ is
  \begin{equation}
    p(\ccsxp | \ccs) = {\binom{|\ccsx| - 1}{\ccs}}^{-1}
  \end{equation}
  Expanding out the RHS of \cref{eq:shapley-value:expectation-equality} we get
  \begin{align}
    &\sum_{\ccs = 0}^{|\ccsx| - 1}p(\ccs)\sum_{\substack{\ccsxp \subseteq \ccsx \setminus \{\ccxi\}\\|\ccsxp| = \ccs}}p(\ccsxp | \ccs) [\ccf(\ccsxp \cup \{\ccxi\}) - \ccf(\ccsxp)] \\
   =& \sum_{\ccs = 0}^{|\ccsx| - 1}\frac{1}{|\ccsx|} \sum_{\substack{\ccsxp \subseteq \ccsx \setminus \{\ccxi\}\\|\ccsxp| = \ccs}}{\binom{|\ccsx| - 1}{\ccs}}^{-1}
[\ccf(\ccsxp \cup \{\ccxi\}) - \ccf(\ccsxp)] \\
   =& \sum_{\ccsxp \subseteq \ccsx \setminus \{\ccxi\}} \frac{1}{|\ccsx|}{\binom{|\ccsx| - 1}{|\ccsxp|}}^{-1}[\ccf(\ccsxp \cup \{\ccxi\}) - \ccf(\ccsxp)] \\
   =& \expectation{\ccsxp}{\ccf(\ccsxp \cup \{\ccxi\}) - \ccf(\ccsxp)}
  \end{align}
\end{proof}

\subsection{Recursive Definition of Variable-Length Input Model}
\label{appendix:recursive-definition-of-aggregation-module}
We define our multi-scale model $\ccf$ (note this is the same as $\ccfms$, but we drop the superscript in this proof for notational simplicity) as a combination of the results from a set of single scale models $\{\ccf^\ccs\}_{\ccs = 1}^{\nmax{}}$
\begin{equation}
  \ccf(\ccsx) = \expectation{\ccrvs}{\expectation{\ccrvxp | \ccrvs}{\ccf^\ccs(\ccrvxp)}}\,.
  \tag{\ref{eq:multi-scale-model} revisited}
\end{equation}
To improve efficiency when computing Shapley values, it is desirable to formulate this in a recursive fashion, which we denote $\check\ccf$.
This enables the computation of $\ccf(\ccsx)$ in terms of the expected result of $\check\ccf(\check{\ccrvx})$ where $\check{\ccrvx}$ is a random variable over subsequences of $\ccsx$ with one element less (all equally probable).
\begin{equation}
  \check\ccf(\ccsx) = \begin{cases}
    \ccf^1(\ccsx) & |\ccsx| = 1 \\
    |X|^{-1} \left[ f^{|X|}(X) + (|X| - 1) \expectation{\check{\ccrvx}}{\check\ccf(\check{\ccrvx})}\right] & |\ccsx| \leq \nmax{} \\
    \expectation{\check{\ccrvx}}{\check\ccf(\check{\ccrvx})} & |\ccsx| > \nmax{} \\
  \end{cases}
  \label{eq:apx:multiscale-model:recursive}
\end{equation}

\begin{proof}
  We prove the equivalence between $\ccf$ and $\check\ccf$ by induction on $|\ccsx|$.

  \hfill \break
  \noindent \textit{Base case:}
  When $|\ccsx| = 1$, observe that $p(\ccs = 1) = 1$ and the sample space $\Omega(\ccrvxp | \ccs = 1) = \{\ccsx\}$ therefore $\ccf(X) = \ccf^1(\ccsx)$ from \cref{eq:multi-scale-model}.

  \hfill \break
  \noindent \textit{Inductive step:}
  We split the inductive step into two parts, one where $|\ccsx| \leq \nmax{}$ and one where $|\ccsx| > \nmax{}$.
  For both parts of the proof, we start by assuming $\ccf(\check{\ccrvx}) = \check\ccf(\check{\ccrvx})$, which we have proven for the base case $s = 1$.
  For $|\ccsx| \leq \nmax{}$, we start from $\check\ccf(\ccsx)$, expanding out the definition according to clause 2 in \cref{eq:apx:multiscale-model:recursive}
  \begin{align}
    \check\ccf(\ccsx) = |X|^{-1} \left[f^{|X|}(X) + (|X| - 1) \expectation{\check{\ccrvx}}{\check\ccf(\check{\ccrvx})} \right]\,.
    \label{eq:apx:step-2}
  \end{align}
  Our strategy will be to expand the expectation in \cref{eq:apx:step-2}, and show that substituting the expanded form back into \cref{eq:apx:step-2} recovers \cref{eq:multi-scale-model}.
  Focusing on the expectation and substituting our assumption, $\ccf(\check{\ccrvx}) = \check\ccf(\check{\ccrvx})$, yields
  \begin{align}
    \expectation{\check{\ccrvx}}{\check\ccf(\check{\ccrvx})} &= \expectation{\check{\ccrvx}}{\ccf(\check{\ccrvx})} \\
    &=\expectation{\check{\ccrvx}}{\expectation{\ccrvs}{\expectation{\check{\ccrvxp} | \ccrvs}{\ccf^\ccs(\ccrvxp)}}}
  \end{align}
  It is important to note the definitions of the random variables here: $\check{\ccrvxp} \subseteq \check{\ccrvx}$ and $\ccs \in \intrange{1}{|\check{\ccrvx}|}$.
  We then expand out the expectations
  \begin{align}
    \expectation{\check{\ccrvx}}{\check\ccf(\check{\ccrvx})} &= \sum_{\substack{\check{\ccrvx} \subset \ccsx\\|\check{\ccrvx}| = |\ccsx| - 1}} \frac{1}{|\ccsx|} \sum_{\ccs = 1}^{|\check{\ccrvx}|} \frac{1}{|\check{\ccrvx}|}\sum_{\substack{\check{\ccrvxp} \subseteq \check{\ccrvx}\\|\check{\ccrvxp}| = \ccs}} {\binom{|\check{\ccrvx}|}{\ccs}}^{-1} \ccf^\ccs(\check{\ccrvxp})\,,
  \end{align}
  reordering the summations, and replacing $|\check{\ccrvx}|$ with $|\ccsx| - 1$ yields
  \begin{align}
    \expectation{\check{\ccrvx}}{\check\ccf(\check{\ccrvx})} &= \frac{1}{|\ccsx|} \sum_{\ccs = 1}^{|\ccsx| - 1} \frac{1}{|\ccsx| - 1}\sum_{\substack{\check{\ccrvx} \subset \ccsx\\|\check{\ccrvx}| = |\ccsx| - 1}}  \sum_{\substack{\check{\ccrvxp} \subseteq \check{\ccrvx}\\|\check{\ccrvxp}| = \ccs}} {\binom{|\ccsx| - 1}{\ccs}}^{-1} \ccf^\ccs(\check{\ccrvxp})\,.
  \end{align}
  As $\{\check{\ccrvx} \subset \ccsx : |\check{\ccrvx}| = |\ccsx| - 1 \}$ is all subsequences of $\ccsx$ excluding a single element, we can substitute this set with $\{\ccxi \in \ccsx\}$ and replace $\check{\ccrvx}$ with $\ccsx \setminus \{\ccxi\}$,
  \begin{align}
    \expectation{\check{\ccrvx}}{\check\ccf(\check{\ccrvx})} &= \frac{1}{|\ccsx|} \sum_{\ccs = 1}^{|\ccsx| - 1} \frac{1}{|\ccsx| - 1} \sum_{\ccxi \in \ccsx}  \sum_{\substack{\ccsxp \subseteq \ccsx \setminus \{\ccxi\}\\|\ccsxp| = \ccs}} {\binom{|\ccsx| - 1}{\ccs}}^{-1} \ccf^\ccs(\ccsxp)\,.
    \label{eq:apx:step-3}
  \end{align}
  Next, we show that
  \begin{align}
    \frac{1}{|\ccsx|} \sum_{\ccxi \in \ccsx}  \sum_{\substack{\ccsxp \subseteq \ccsx \setminus \{\ccxi\}\\|\ccsxp| = \ccs}} {\binom{|\ccsx| - 1}{\ccs}}^{-1}
    = \sum_{\substack{\ccsxp \subseteq \ccsx\\|\ccsxp| = \ccs}} {\binom{|\ccsx|}{\ccs}}^{-1}\,
    \label{eq:apx:combining-summations}
  \end{align}
  To get from the LHS to the RHS, we need to consider how many times we will repeatedly count $\ccsxp$ due to the outer summation.
  If we consider a fixed $\ccsxp$, then there will be $|\ccsx| - |\ccsxp|$ times where it will be drawn in the inner summation.
  Hence we can replace the outer summation with the number of occurrences
  \begin{align}
    &\frac{1}{|\ccsx|} \sum_{\ccxi \in \ccsx}  \sum_{\substack{\ccsxp \subseteq \ccsx \setminus \{\ccxi\}\\|\ccsxp| = \ccs}} {\binom{|\ccsx| - 1}{\ccs}}^{-1} \\
    =& \frac{1}{|\ccsx|} \sum_{\substack{\ccsxp \subseteq \ccsx \setminus \{\ccxi\}\\|\ccsxp| = \ccs}} (|\ccsx| - \ccs){\binom{|\ccsx| - 1}{\ccs}}^{-1} \\
    =&  \sum_{\substack{\ccsxp \subseteq \ccsx \setminus \{\ccxi\}\\|\ccsxp| = \ccs}} \frac{(|\ccsx| - \ccs)}{|\ccsx|} \frac{(|\ccsx| - 1 - \ccs)!\ccs!}{(|\ccsx| - 1)!} \\
    =&  \sum_{\substack{\ccsxp \subseteq \ccsx \setminus \{\ccxi\}\\|\ccsxp| = \ccs}} \frac{(|\ccsx| - \ccs)!\ccs!}{(|\ccsx|)!} \\
    =&  \sum_{\substack{\ccsxp \subseteq \ccsx \setminus \{\ccxi\}\\|\ccsxp| = \ccs}} {\binom{|\ccsx|}{\ccs}}^{-1}\,.
  \end{align}
  Having proven \cref{eq:apx:combining-summations}, we can simplify \cref{eq:apx:step-3} to
  \begin{align}
    \expectation{\check{\ccrvx}}{\check\ccf(\check{\ccrvx})} &= \frac{1}{|\ccsx| - 1} \sum_{\ccs = 1}^{|\ccsx| - 1}   \sum_{\substack{\ccsxp \subseteq \ccsx\\|\ccsxp| = \ccs}} {\binom{|\ccsx|}{\ccs}}^{-1} \ccf^\ccs(\ccsxp)\,.
  \end{align}
  Substituting this form back into \cref{eq:apx:step-2} we get
  \begin{align}
    \check\ccf(\ccsx) = |X|^{-1} \left[ \sum_{\ccs = 1}^{|\ccsx| - 1}   \sum_{\substack{\ccsxp \subseteq \ccsx\\|\ccsxp| = \ccs}} \left[ {\binom{|\ccsx|}{\ccs}}^{-1} \ccf^\ccs(\ccsxp)  \right]+ f^{|X|}(X)\right].
  \end{align}
  Now $f^{|X|}(X)$ can be merged into the summation over $\ccs$ by increasing its bound from $|\ccsx| - 1$ to $|\ccsx|$
  \begin{align}
    \check\ccf(\ccsx) =   \sum_{\ccs = 1}^{|\ccsx|}  \frac{1}{|X|} \sum_{\substack{\ccsxp \subseteq \ccsx\\|\ccsxp| = \ccs}}  {\binom{|\ccsx|}{\ccs}}^{-1} \ccf^\ccs(\ccsxp)\,.
  \end{align}
  Which is the expansion of the expectations of $\ccf(\ccsx)$.
  The proof for the equivalence between $\check\ccf(\ccsx)$ and $\ccf(\ccsx)$ for $|\ccsx| > \nmax{}$ is very similar to the above and has been omitted for brevity.
\end{proof}

\subsection{Linearity of Shapley Values}
\label{appendix:linearity-of-shapley-values}
The Shapley value $\phi^{w_1c_1 + w_2c_2}_i$ for a model $f_{w_1c_1 + w_2c_2}(X) = w_1 f_{c_1}(X) + w_2 f_{c_2}(X)$ where $w_1, w_2 \in \mathbb{R}$ is $w_1\phi^{c_1}_i + w_2\phi^{c_2}_i$.
\begin{proof}
\begin{align}
\phi^{w_1c_1 + w_2c_2} &= \sum_{\ccsxp \subseteq \ccsx \setminus \{\ccxi\}} w(X') f_{w_1c_1 + w_2c_2}(\ccsxp) \\
                 &= \sum_{\ccsxp \subseteq \ccsx \setminus \{\ccxi\}} w(X') \left[ w_1 f_{c_1}(\ccsxp) + w_2 f_{c_2}(\ccsxp) \right] \\
                 &= w_1 \sum_{\ccsxp \subseteq \ccsx \setminus \{\ccxi\}} w(X') f_{c_1}(\ccsxp) +  w_2 \sum_{\ccsxp \subseteq \ccsx \setminus \{\ccxi\}} w(X') f_{c_2}(\ccsxp) \\
                 &= w_1 \phi^{c_1}_i +  w_2 \phi^{c_2}_i
\end{align}

\end{proof}

\section{Detailed Experimental Setup}
\label{appendix:experimental-setup}

\paragraph{TRN.}
We extract 256D features for every frame in Something-something v2 using the publicly available 8 frame multi-scale TRN\footnote{\url{https://github.com/zhoubolei/TRN-pytorch}} (this uses a BN-Inception~\cite{ioffe2015_BatchNormalizationAccelerating} backbone).
We then train a set of MLP classifiers $\{\ccf^\ccs\}^{\nmax{} = 8}_{\ccs = 1}$ with one hidden layer of 256 units, dropout of 0.1 on the input features, and ReLU activation.
Each MLP $\ccf^\ccs$ takes in the concatenation of $\ccs$ frame features and produces class scores.
This is the same as the multi-scale variant of TRN, however, rather than training the classifiers jointly, we train them separately and then combine their results at inference time through \cref{eq:multi-scale-model}.
We tried training these classifiers jointly, producing a final class score by \cref{eq:multi-scale-model}, however the individual classifiers perform very poorly when tested in isolation, and the overall performance was improved by training them separately.
We train for 30 epochs with a batch-size of 512 and learning rate of 1e-3 using Adam to optimise parameters.

\paragraph{TSN.}
We train TSN~\cite{wang2016_TemporalSegmentNetworks} with a ResNet-50 backbone~\cite{he2016_DeepResidualLearning}.
We introduce a bottleneck FC layer mapping from the 2048D features from the GAP layer to 256D, we then add a FC layer atop of this for classification.
The Element Shapley Values for TSN can be obtained since the model is inherently capable of operating on variable length sequences.
This means we can directly evaluate $\ccfc(\ccsxp)$ for all subsequences $\ccsxp$.
When scaling up to a large number of frames, we use the approximation technique as specified in \cref{alg:shapley-value-computation} but lines 7-12 are replaced with
\begin{equation}
    \mathcal{F}^s_j \leftarrow f_c(\mathcal{X}^s_j)\,.
\end{equation}
This adaptation to the algorithm is the same for any other model already supporting a variable length input.

\begin{figure}[htb]
  \centering
  \includegraphics[width=.4\textwidth]{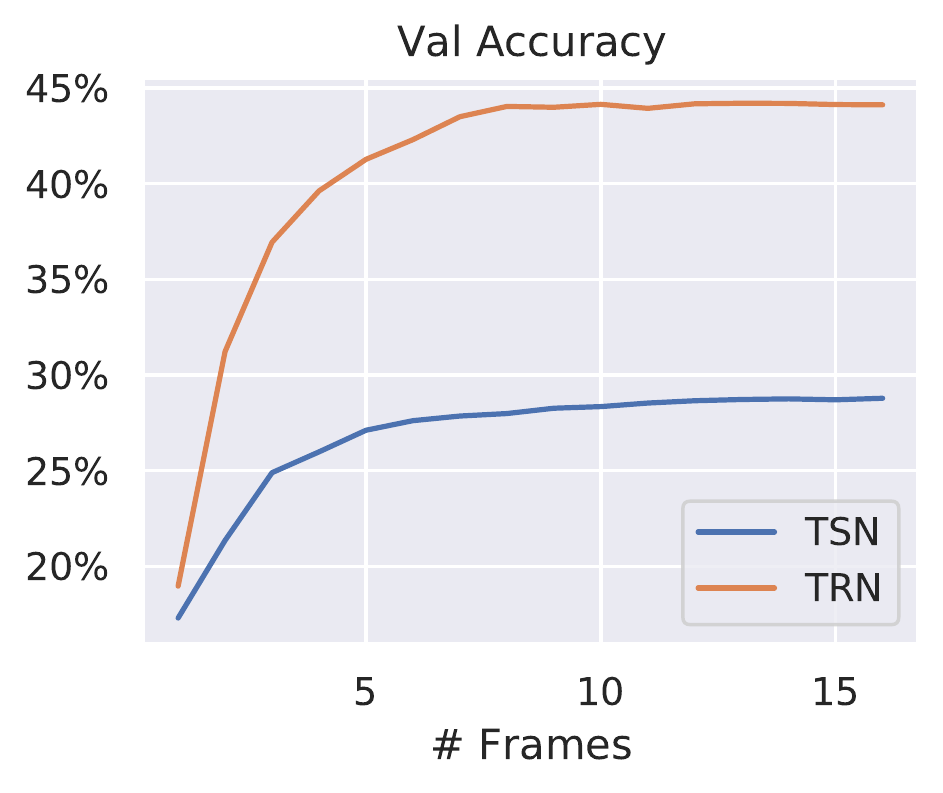}
  \caption{TRN and TSN validation accuracy by number of frames input to the model.}
  \label{fig:model-accuracy-by-frame-counts}
\end{figure}

The accuracy of both models on the validation set is shown in \cref{fig:model-accuracy-by-frame-counts} across a range of different sequence lengths where frames are uniformly sampled from the video.

\section{Computational Cost}
\label{appendix:computationl-cost}
The computational cost of the Element Shapley Value without approximation is $\mathcal{O}(2^{|X|})$ in the number of forward passes.
When we apply our approximation method presented in \cref{alg:shapley-value-computation} we reduce that complexity to $\mathcal{O}(mi|X|)$ where $m$ is the maximum number of sequences sampled per scale and $i$ is the number of iterations of the approximation.
Practically speaking, we can compute Element Shapley Values for both TSN and TRN exactly in reasonable time (under ~10s/example) thanks to an optimised batched GPU implementation for up to 16 frames.
Beyond this, the exponential scaling results in prohibitively long runtimes.
Our approximation enables us to scale to more than 16 frames.
We present a runtime analysis on a single NVIDIA 1080Ti in \cref{tab:runtime}.

\begin{table}[ht]
\label{tab:runtime}
\caption{Runtime analysis of exact and approximate ESV on sequences of varying
  length.
  When applying our approximation algorithm
  (\cref{alg:shapley-value-computation}) we have to choose $m$ (max number of subsequences/scale) and $i$ (number of approximation iterations).}
  \centering
\begin{tabular}{@{}l@{\hskip .2in}l@{\hskip .2in}l@{\hskip .2in}l@{}}
\toprule
\multicolumn{1}{l}{Model} & No. of frames & Configuration   & Time (s)           \\ \midrule
\multirow{3}{*}{TRN}      & 8             & Exact           & $.080\pm0.0001$   \\
                          & 16            & Exact           & $6.79\pm0.01$      \\
                          & 16            & m = 1024, i = 1 & $.122\pm0.005$    \\\midrule
\multirow{4}{*}{TSN}      & 8             & Exact           & $.0043\pm0.0007$ \\
                          & 16            & Exact           & $.035\pm0.0001$   \\
                          & 20            & Exact           & $.614\pm0.006$    \\
                          & 20            & m = 1024, i = 1 & $.116\pm0.0005$   \\ \bottomrule
\end{tabular}
\end{table}





\section{Additional Analysis for Shapley Values using TSN}
\label{appendix:tsn-results}

\begin{figure}[htb]
  \centering
  \begin{minipage}[t]{0.375\textwidth}
    \vspace{0pt}  
    \begin{overpic}[width=\textwidth]{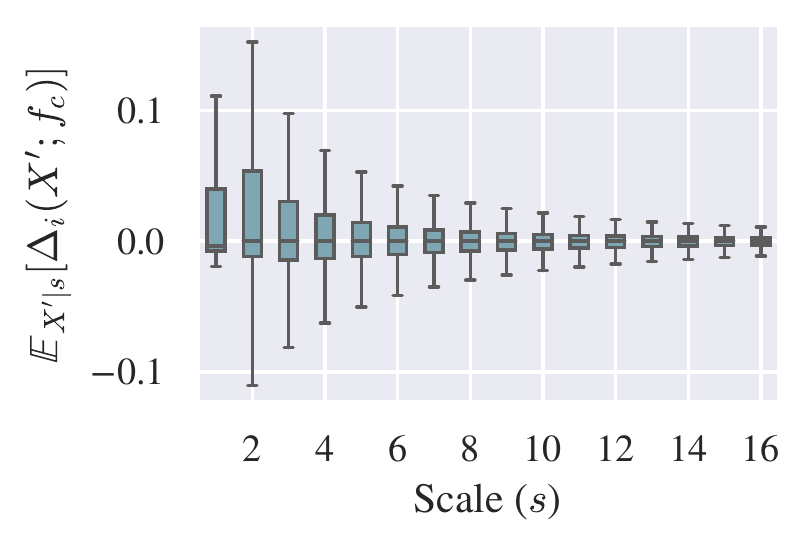}
    \put (2, 8) {(a)}
    \end{overpic}
    \phantomsubcaption
    \label{fig:average-marginal-contributions-across-scales:tsn}
  \end{minipage}
   \begin{minipage}[t]{0.27\textwidth}
    \vspace{0pt}  
    \begin{overpic}[width=\textwidth]{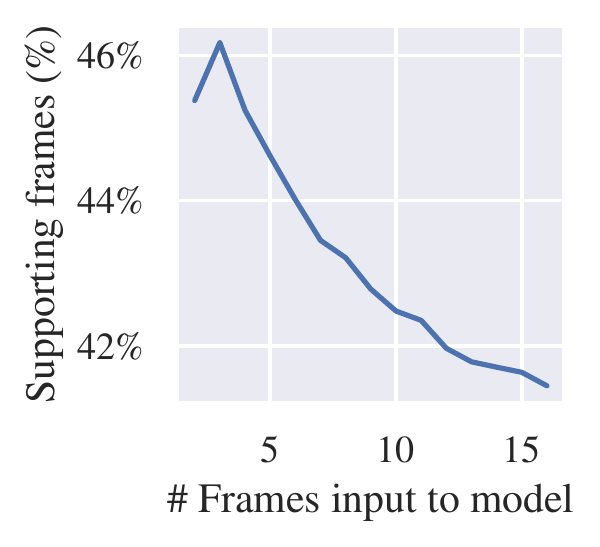}
    \put (2, 3) {(b)}
    \end{overpic}
    \phantomsubcaption
    \label{fig:shapley-value:varying-number-of-input-frames:tsn}
  \end{minipage}
  \begin{minipage}[t]{0.265\textwidth}
    \vspace{0pt}  
    \begin{overpic}[width=\textwidth]{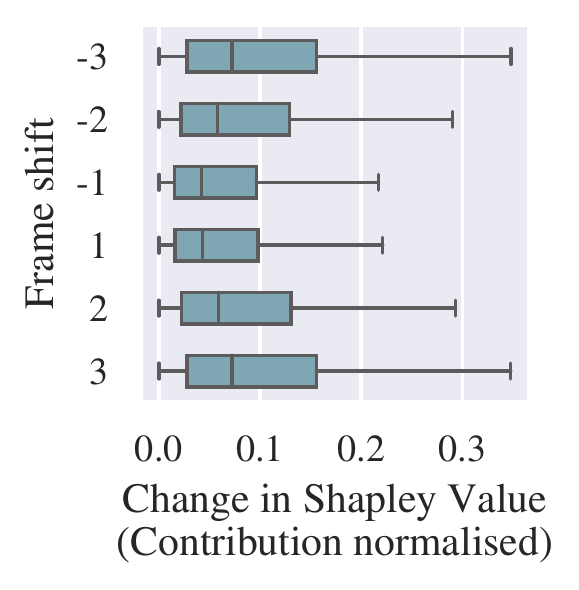}
    \put (2, 10) {(c)}
    \end{overpic}
    \phantomsubcaption
    \label{fig:temporal-stability:tsn}
  \end{minipage}

  \caption{%
    TSN: \subref{fig:average-marginal-contributions-across-scales:tsn}
    Box-plot of marginal contributions at each scale.
    \subref{fig:shapley-value:varying-number-of-input-frames:tsn} Increasing the number of frames fed to the model decreases the percentage of supporting frames.
    \subref{fig:temporal-stability:tsn}~Change in ESV when compared to neighbouring frames. Larger shifts result in larger differences in ESV indicating temporal smoothness.
    (cf. \cref{fig:scale-analysis:trn} for TRN).
  }
  \label{fig:scale-analysis:tsn}
\end{figure}

In this section we present analogous results for TSN, compared to the TRN counterparts in the main paper.
\Cref{fig:average-marginal-contributions-across-scales:tsn} reports the average marginal contribution at each scale (cf.\ \cref{fig:average-marginal-contributions-across-scales:trn}).
The trends seen here are much the same as for TRN: as the scale increases, the average marginal contribution decreases.
\Cref{fig:shapley-value:varying-number-of-input-frames:tsn} shows the number of supporting frames vs.\ number of frames input to the model (cf.\ \cref{fig:shapley-value:varying-number-of-input-frames:trn}).
Whilst the trend is similar to that for TRN, the proportion of supporting frames for TSN is much lower.
\Cref{fig:temporal-stability:tsn} shows the results of the frame shifting experiment for TSN (cf.\ \Cref{fig:temporal-stability:trn}) also demonstrating temporal smoothness.

\begin{figure}[t]
  \centering
  \begin{minipage}[t]{\textwidth}
    \vspace{0pt}  
    \centering
    \begin{overpic}[width=\textwidth]{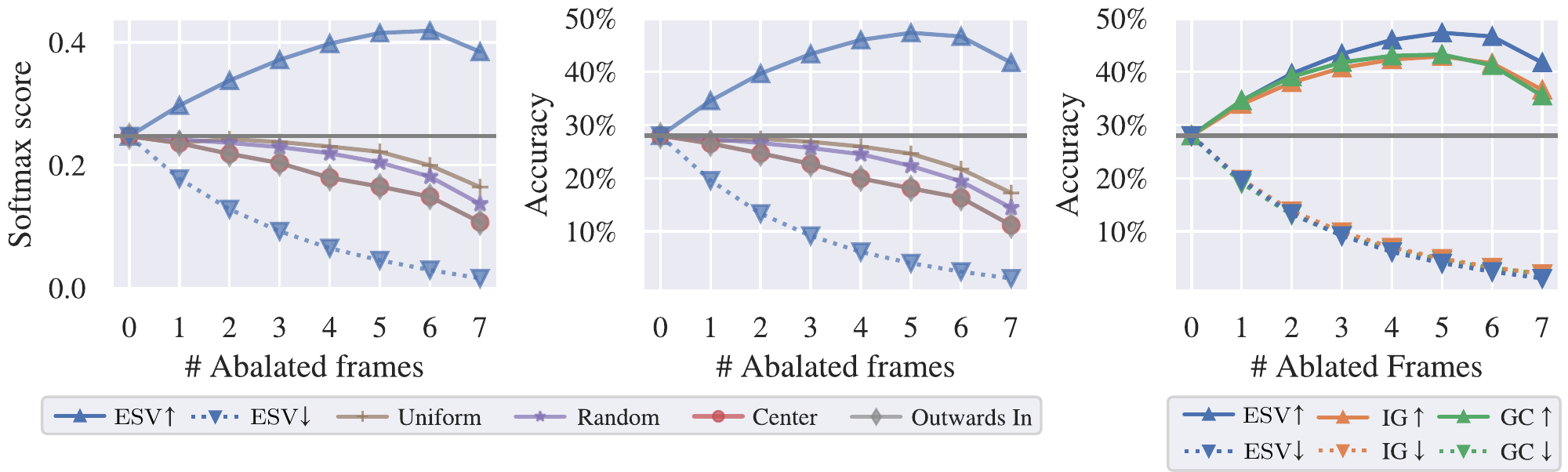}
    \put (0, 6) {(a)}
    \put (33.5,6) {(b)}
    \put (67.5,6) {(c)}
    \end{overpic}
    \phantomsubcaption
    \label{fig:frame-ablation-tsn:score}
    \phantomsubcaption
    \label{fig:frame-ablation-tsn:accuracy}
    \phantomsubcaption
    \label{fig:method-comparison-frame-ablation-tsn}
  \end{minipage}
  \vspace{-2em}
  \caption{
    TSN class score and accuracy after iteratively discarding frames in order of their attribution rank (ascending $\blacktriangle$ vs descending $\blacktriangledown$).
    We compare our method~(ESV) to baselines (a,b) and two alternate attribution methods in (c): GradCam (GC) and Integrated Gradients (IG), keeping figures (b) and (c) separate for legibility.
    (cf.\ \cref{fig:frame-ablation} for TRN).
    }
  \label{fig:frame-ablation:tsn}
\end{figure}

We conduct the same experiment on frame discarding for TSN as shown for TRN in \cref{fig:frame-ablation}, presenting results in \cref{fig:frame-ablation:tsn}.
Since both GC and IG produce similar attribution values to ESV for TSN, we see less of a performance gap when discarding frames by the attribution ranks in descending order.
 However, ESV still produces a larger performance gain than GC and IG when removing frames in ascending order of attributions.

\begin{figure}[t]
  \centering
  \begin{minipage}[t]{.4\textwidth}
    \centering
    TRN
  \end{minipage}
  \begin{minipage}[t]{.4\textwidth}
    \centering
    TSN
  \end{minipage}
  \begin{minipage}[t]{.8\textwidth}
    \includegraphics[width=.5\textwidth]{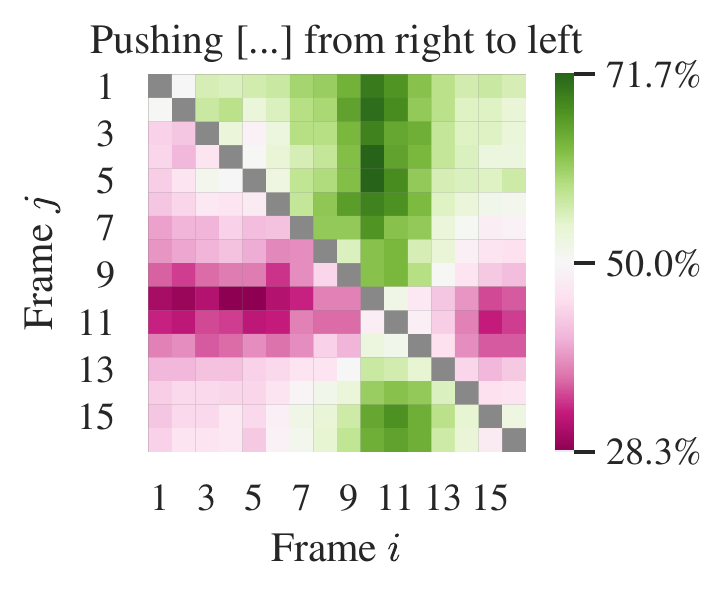}
    \includegraphics[width=.5\textwidth]{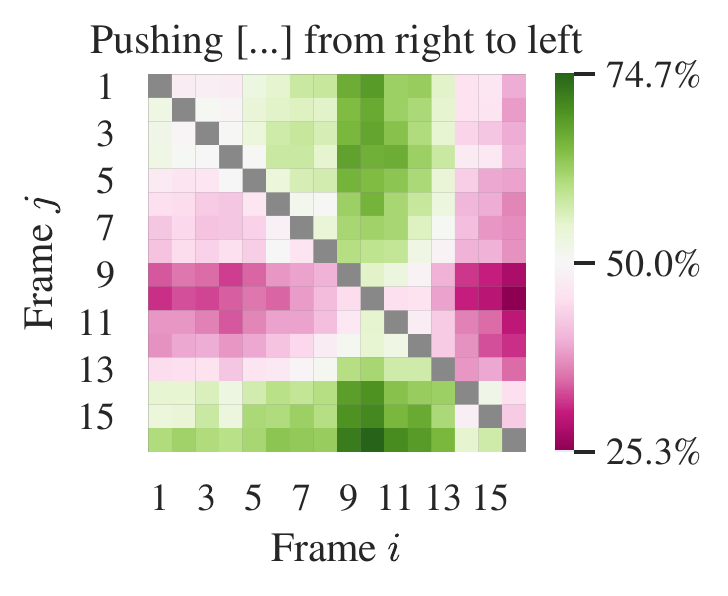}
  \end{minipage}
  \begin{minipage}[t]{.8\textwidth}
    \includegraphics[width=.5\textwidth]{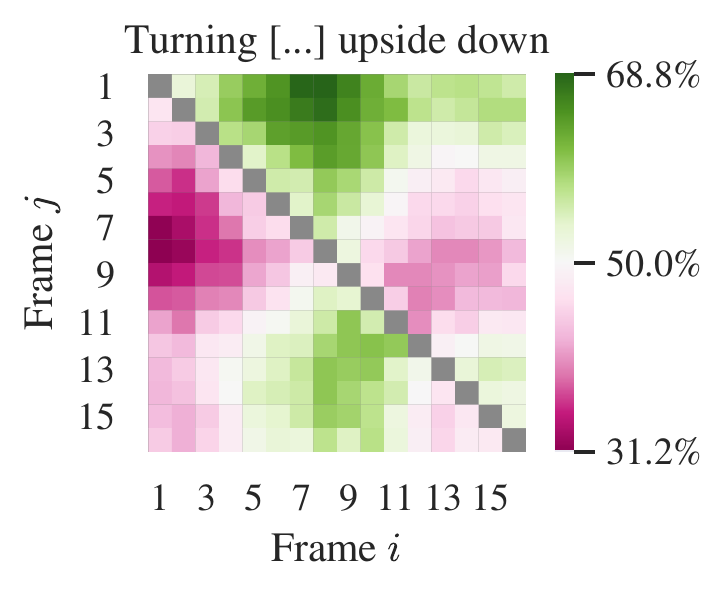}
    \includegraphics[width=.5\textwidth]{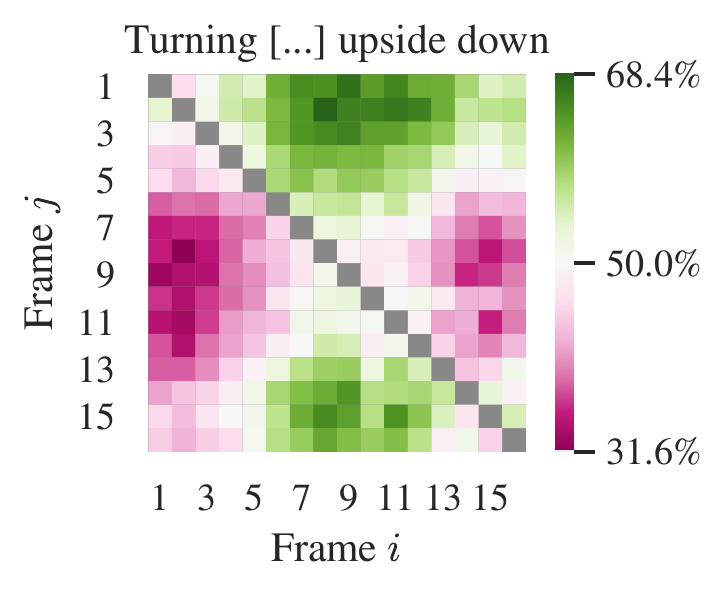}
  \end{minipage}
  \begin{minipage}[t]{.8\textwidth}
    \includegraphics[width=.5\textwidth]{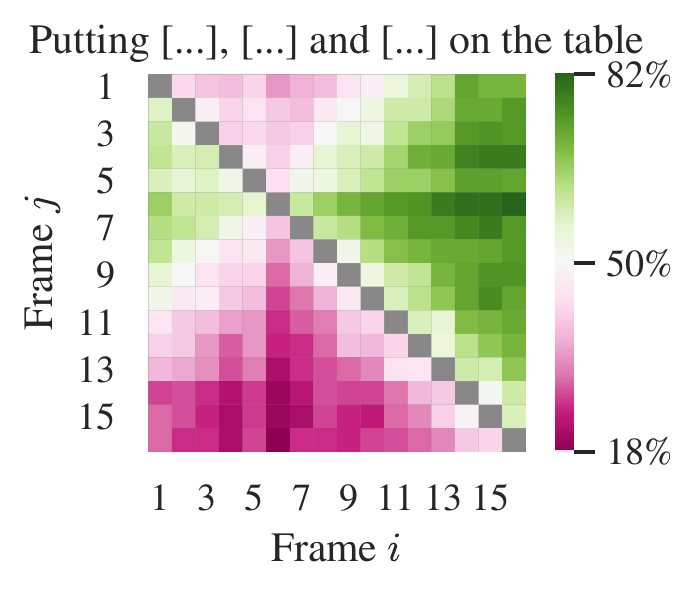}
    \includegraphics[width=.5\textwidth]{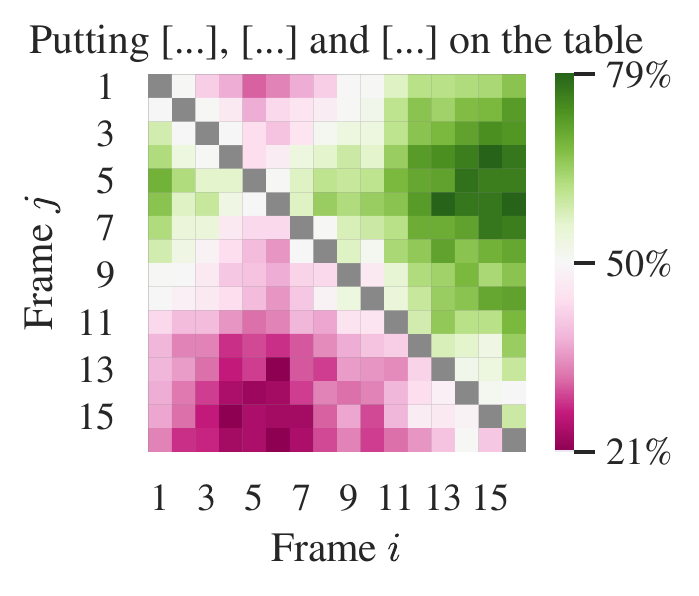}
  \end{minipage}
  \caption{Comparing the percentage of videos where $\phi^c_\cci > \phi^c_\ccj$ for TRN (left) and TSN (right). These are classes where the models value frames from the same position similarly.}
  \label{fig:phi_i_gt_phi_j:similar}
\end{figure}

\begin{figure}[t]
  \centering
  \begin{minipage}[t]{.4\textwidth}
    \centering
    TRN
  \end{minipage}
  \begin{minipage}[t]{.4\textwidth}
    \centering
    TSN
  \end{minipage}
  \begin{minipage}[t]{.8\textwidth}
    \includegraphics[width=.5\textwidth]{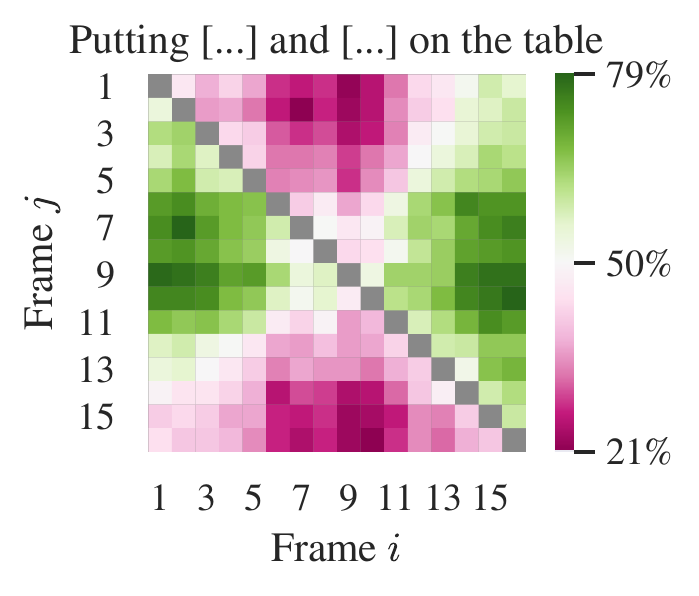}
    \includegraphics[width=.5\textwidth]{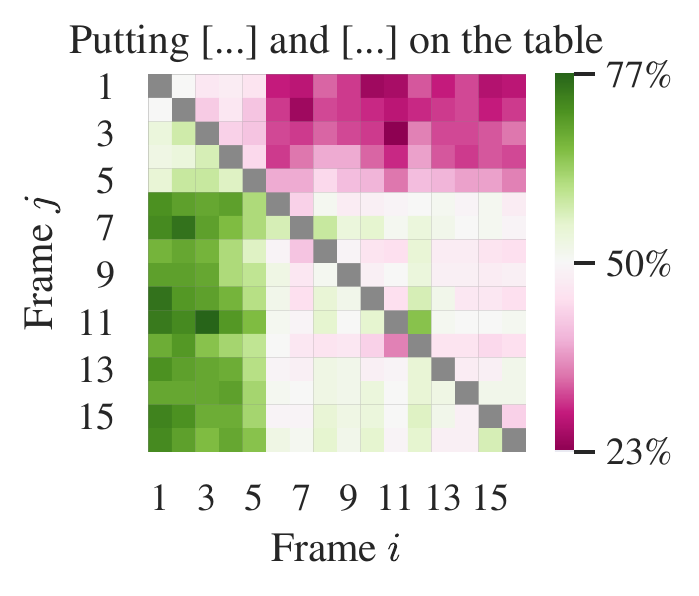}
  \end{minipage}
  \begin{minipage}[t]{.8\textwidth}
    \includegraphics[width=.5\textwidth]{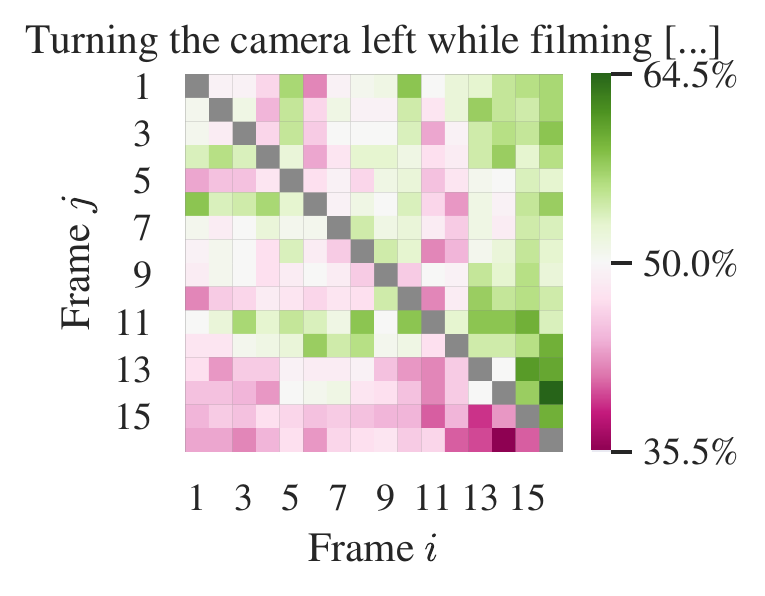}
    \includegraphics[width=.5\textwidth]{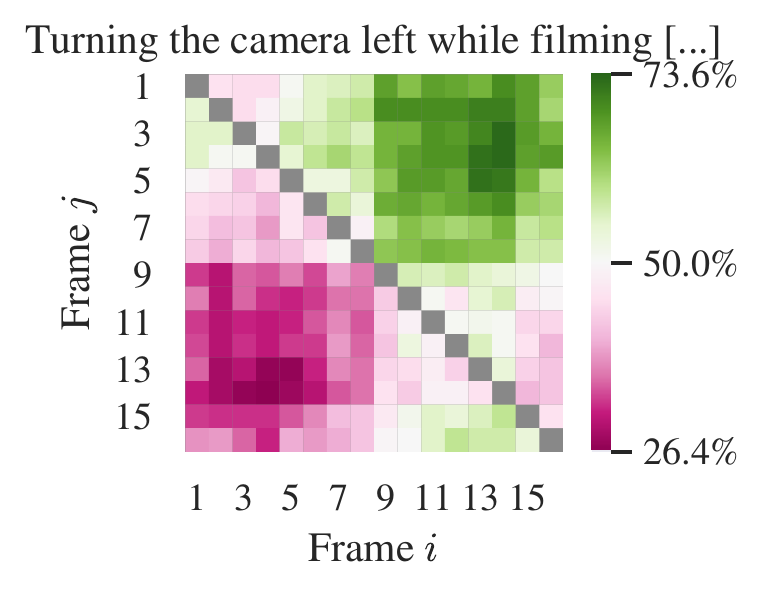}
  \end{minipage}
  \begin{minipage}[t]{.8\textwidth}
    \includegraphics[width=.5\textwidth]{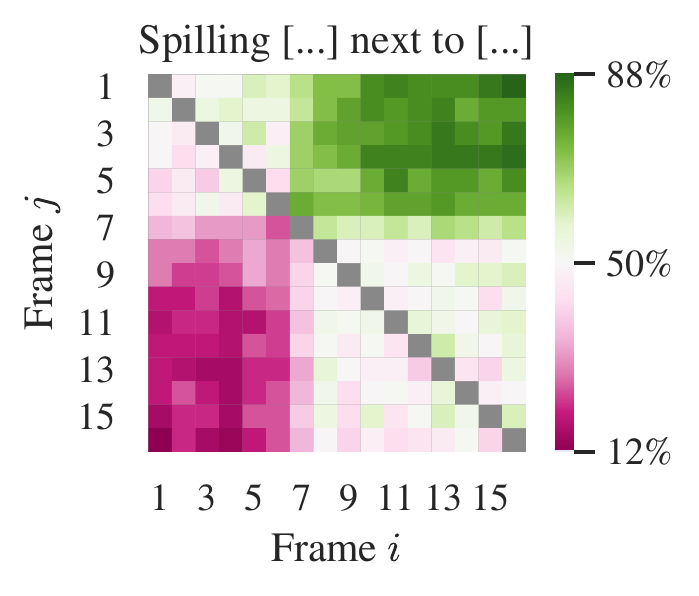}
    \includegraphics[width=.5\textwidth]{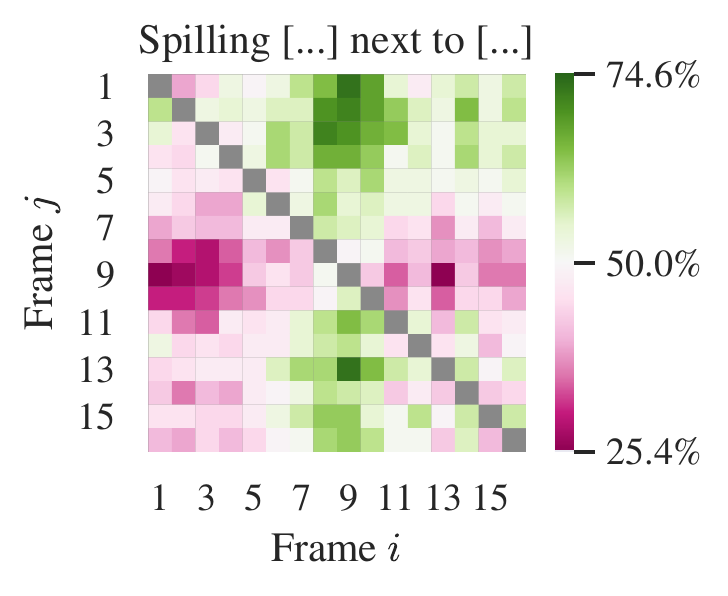}
  \end{minipage}
  \caption{Comparing the percentage of videos where $\phi^c_\cci > \phi^c_\ccj$ for TRN (left) and TSN (right). These are classes where the models value frames from the same position differently.}
  \label{fig:phi_i_gt_phi_j:different}
\end{figure}

Finally, in \cref{fig:phi_i_gt_phi_j:similar,fig:phi_i_gt_phi_j:different}, we present heatmaps showing the percentage of videos where $\phi^c_i > \phi^c_i$ for both TSN and TRN.
\Cref{fig:phi_i_gt_phi_j:similar} shows classes where the distributions are similar and \Cref{fig:phi_i_gt_phi_j:different} where they are different.


\end{document}